\documentclass{article}

\usepackage{iclr2027_conference,times}

\usepackage[utf8]{inputenc}
\usepackage[T1]{fontenc}
\usepackage{graphicx}
\usepackage{wrapfig}
\usepackage{capt-of}
\usepackage{needspace}
\usepackage{hyperref}
\usepackage{url}
\usepackage{booktabs}
\usepackage{multirow}
\usepackage{amsfonts}
\usepackage{nicefrac}
\usepackage{microtype}
\usepackage{xcolor}
\usepackage{colortbl}
\usepackage{xspace}
\usepackage{amsmath}
\usepackage{amssymb}
\usepackage{bm}
\usepackage{relsize}
\usepackage{tcolorbox}
\usepackage{tikz}
\newsavebox{\maintablebox}
\newcommand{\maintablescale}{1}

\newcommand{\SEthree}{\ensuremath{\mathrm{SE}(3)}\xspace}
\newcommand{\linkToPdf}[1]{\href{#1}{[pdf]}}
\newcommand{\linkToVideo}[1]{\href{#1}{[video]}}
\newcommand{\linkToCode}[1]{\href{#1}{[code]}}
\newcommand{\linkToWeb}[1]{\href{#1}{[web]}}

\newcommand{\myParagraph}[1]{{\bf #1.}\xspace}

\newcommand{\bestcell}[1]{\cellcolor{green!60}#1}
\newcommand{\secondcell}[1]{\cellcolor{yellow!60}#1}
\newcommand{\runnerupcell}[1]{\cellcolor{yellow!60}#1}

\newtcolorbox[auto counter]{greytheorem}[2][]{
  colback=black!5,
  colframe=black!25,
  coltitle=black,
  fonttitle=\bfseries,
  title={Theorem~\thetcbcounter: #2},
  #1,
  boxrule=0.4pt,
  arc=1pt,
  left=5pt,
  right=5pt,
  top=3pt,
  bottom=3pt,
  before skip=4pt,
  after skip=4pt
}

\newtcolorbox[use counter from=greytheorem]{greydefinition}[2][]{
  colback=black!5,
  colframe=black!25,
  coltitle=black,
  fonttitle=\bfseries,
  title={Definition~\thetcbcounter: #2},
  #1,
  boxrule=0.4pt,
  arc=1pt,
  left=5pt,
  right=5pt,
  top=3pt,
  bottom=3pt,
  before skip=4pt,
  after skip=4pt
}

\title{Matisse: Evidence-Space Reasoning for Active 3D Reconstruction}

\author{%
  \hypersetup{pdfborder={0 0 0}}
\textbf{%
  Xihang Yu$^{1}$\thanks{Corresponding author: \texttt{jimmyyu@mit.edu}. More details can be found at \href{https://xihangyu630.github.io/matisse/}{\textcolor{red}{the project website}}.},
  Kaichen Zhou$^{1}$,
  Lorenzo Shaikewitz$^{1}$,
  Clément Jambon$^{1}$,
  Xiao Zhan$^{1}$} \\
\textbf{Rajat Talak$^{2}$, Luca Carlone$^{1}$} \\[0.5em]
$^{1}$Massachusetts Institute of Technology
\qquad
$^{2}$National University of Singapore

}

\iclrpreprintcopy
\begin{document}

\maketitle

\begin{NoHyper}
\renewcommand{\thefootnote}{}
\footnotetext{%
This work was partially supported by Symbotic, DCIST, the ONR RAPID program,
and AFOSR "Certifiable and Self-Supervised Category-Level Tracking" Program.
L.~Shaikewitz is supported by an NSF graduate research fellowship.
L.~Carlone holds concurrent appointments as a faculty member at the Massachusetts
Institute of Technology and as an Amazon Scholar. This paper describes work
performed at MIT and is not associated with Amazon.}
\end{NoHyper}

\begin{figure}[!h]
  \centering
  \begin{minipage}{1.0\linewidth}
  \centering
  \begin{tikzpicture}[x=\linewidth,y=\linewidth]
    \node[inner sep=0pt,outer sep=0pt] at (0.125,0)
      {\includegraphics[width=0.25\linewidth]{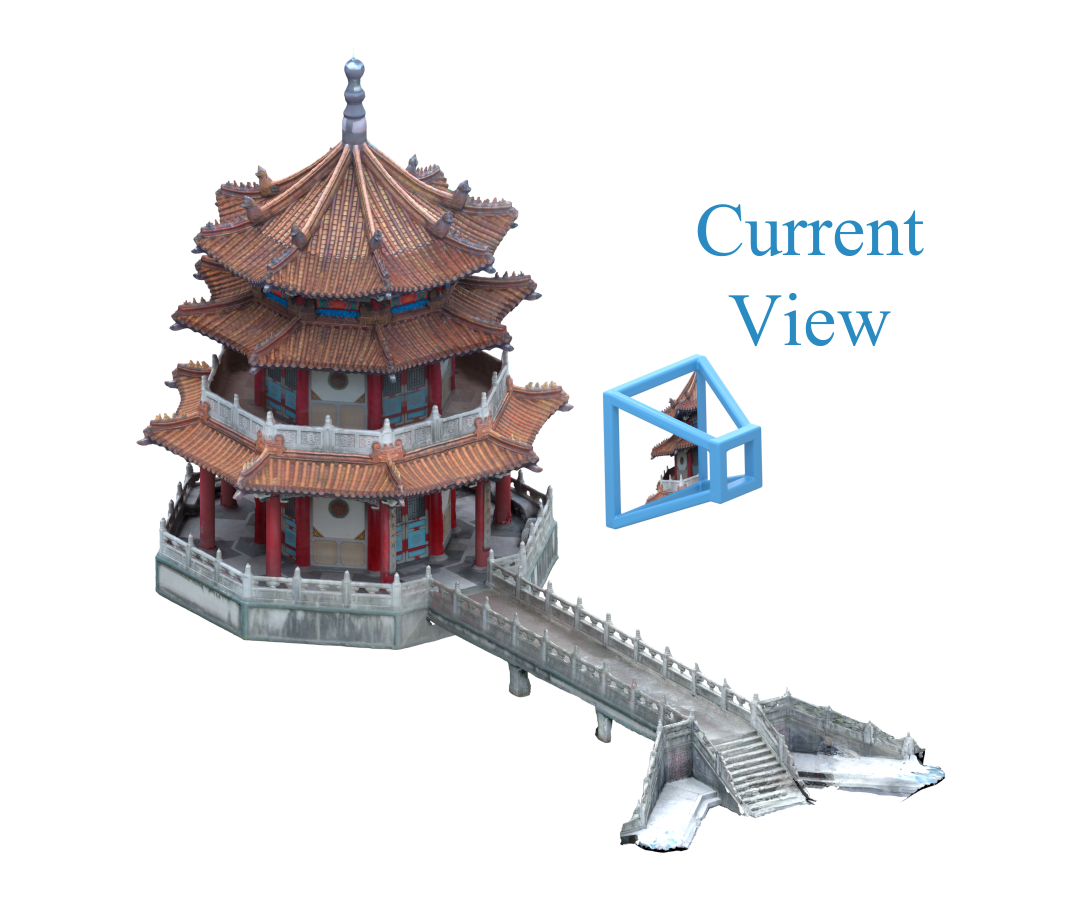}};
    \node[inner sep=0pt,outer sep=0pt] at (0.375,0)
      {\includegraphics[width=0.25\linewidth]{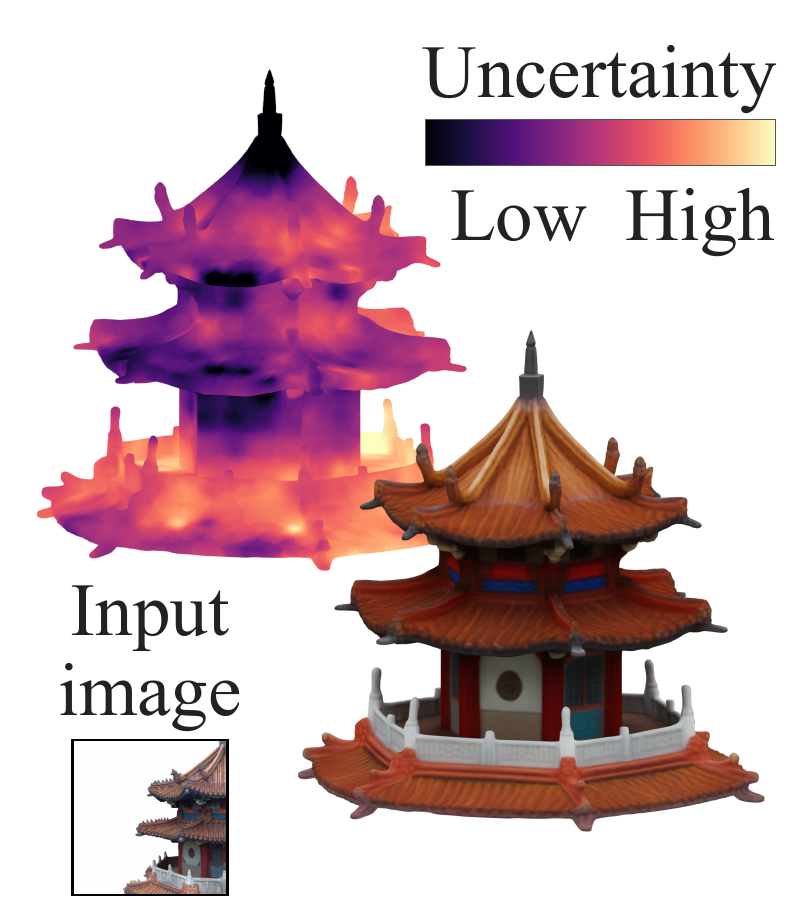}};
    \node[inner sep=0pt,outer sep=0pt] at (0.625,0)
      {\includegraphics[width=0.25\linewidth]{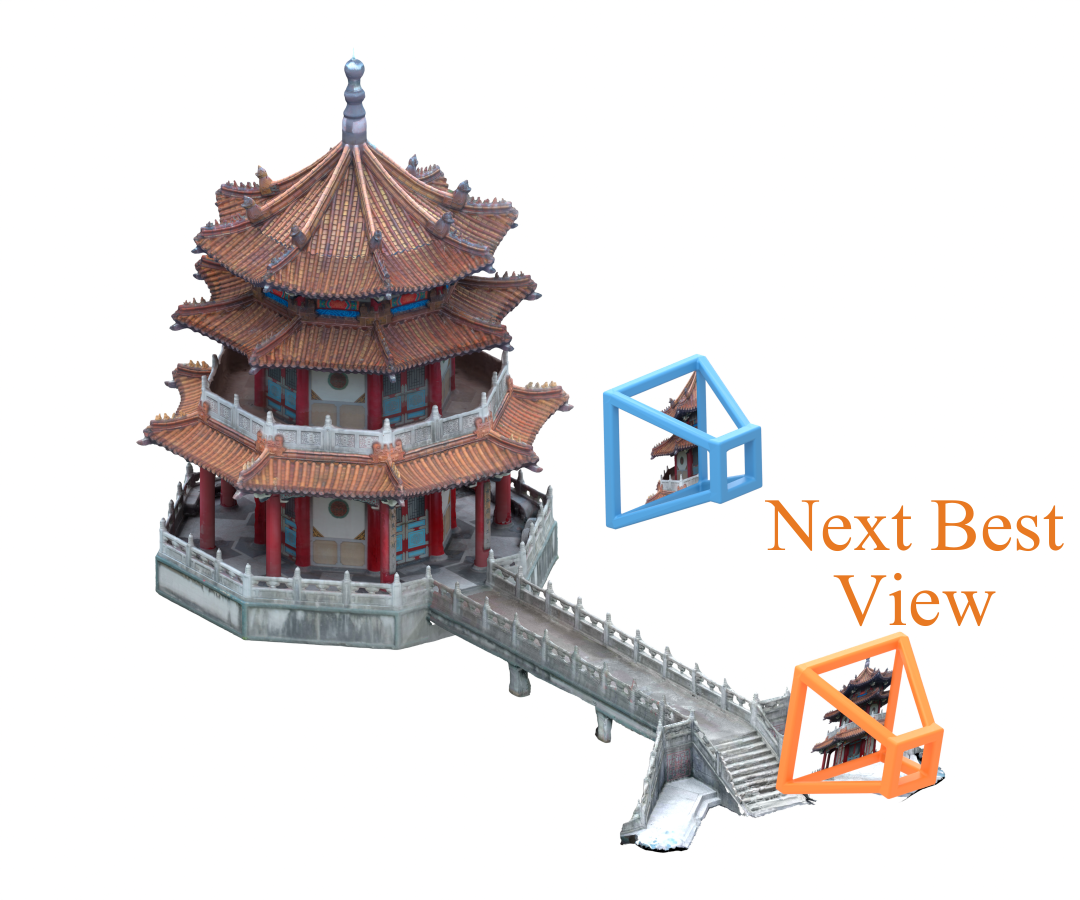}};
    \node[inner sep=0pt,outer sep=0pt] at (0.875,0)
      {\includegraphics[width=0.25\linewidth]{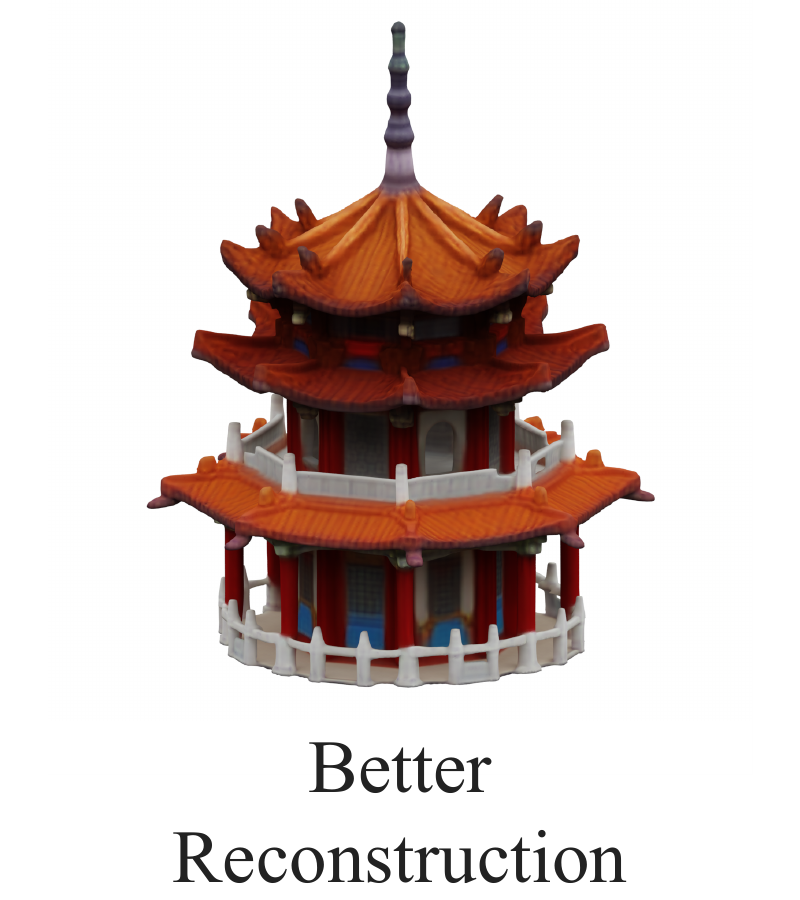}};
    \draw[-stealth,line width=1.5pt] (0.205,0) -- (0.265,0);
    \draw[-stealth,line width=1.5pt] (0.47,0) -- (0.53,0);
    \draw[-stealth,line width=1.5pt] (0.72,0) -- (0.78,0);
    \node[overlay,anchor=north west,xshift=2pt,yshift=-2pt]
      at (current bounding box.north west) {\small\textbf{(a)}};
  \end{tikzpicture}\par
  \vspace{0.5em}
  \begin{tikzpicture}[x=\linewidth,y=\linewidth]
    \node[inner sep=0pt,outer sep=0pt] at (0.5,0)
      {\includegraphics[width=\linewidth]{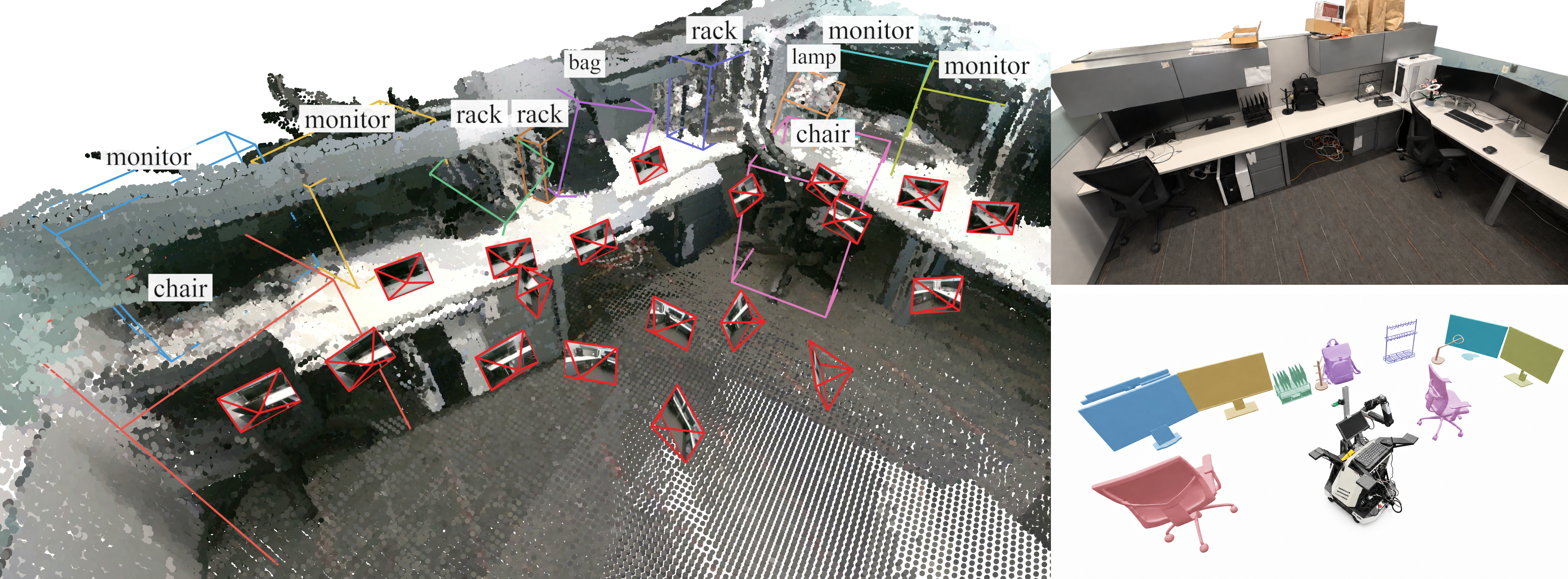}};
    \node[overlay,anchor=north west,xshift=2pt,yshift=-2pt]
      at (current bounding box.north west) {\small\textbf{(b)}};
  \end{tikzpicture}\par
  \end{minipage}
  \caption{Overview of Matisse. \textbf{(a) Active reconstruction workflow.}
  Given the current observation, Matisse
  estimates uncertainty over observed and unobserved geometry, selects the
  next best view, and as more frames are acquired, it decides which keyframes to retain to 
  achieve a more complete reconstruction. \textbf{(b) Real robot experiment.}
  The aggregated depth point cloud with selected camera views and object bounding boxes
  (left), alongside the real captured scene image (upper right) and reconstructed object meshes
  (lower right).}
  \label{fig:matisse-overview}
  \vspace{-0.9em}
\end{figure}
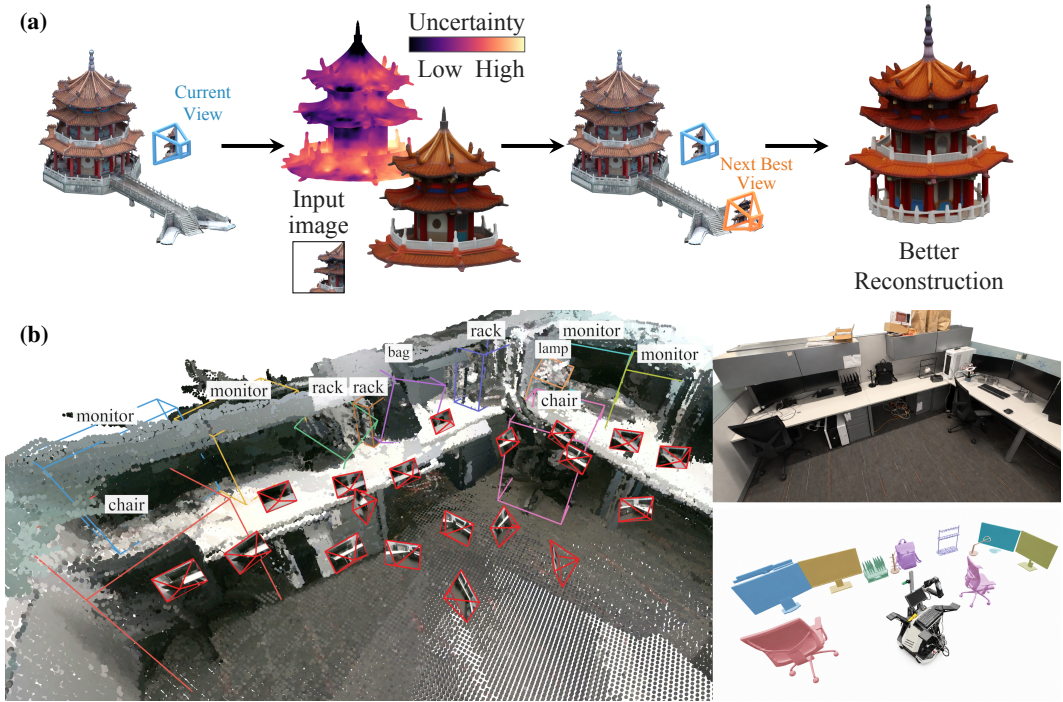

\begin{abstract}
How can a 3D reconstruction system acquire and retain useful information to
understand the geometry of a scene from partial views under a limited computation
budget? Existing active view acquisition methods typically estimate 
uncertainty over observed or instantiated geometry, 
limiting their ability to reason about unseen structure, while long-horizon reconstruction 
methods often retain redundant observations. 
We introduce \emph{Matisse}, a training-free 
framework that unifies active reconstruction and keyframe selection by leveraging
evidence provided by a pretrained generative 3D model. Matisse estimates
Evidential Uncertainty from cross-attention evidence associated with 3D latent
tokens and derives an Evidential Information Gain to guide both
view acquisition and keyframe selection based on the expected reduction in posterior entropy.
Matisse supports multi-object scenes through occlusion-aware, object-balanced 
aggregation and propagates uncertainty through intermediate latents to 
avoid full reconstruction during planning. Matisse reduces
Chamfer distance by $12.7\%$, $3.8\%$, and $9.2\%$ on GSO30, YCB-V,
and Replica, respectively, relative to the best baseline on each dataset,
and achieves a $1.50\times$ end-to-end speedup over the best active
reconstruction baseline on GSO30 with the same backend.
In the GSO30 keyframe selection experiment for long-horizon reconstruction,
Matisse achieves $18.3\%$ lower Chamfer distance
than Stream3D and approaches convergence using roughly $20\%$ as many views.
\end{abstract}


\section{Introduction}
\label{sec:introduction}
How can a 3D reconstruction system 
\citep{chen26cvpr-sam,xiang26cvpr-native,zhang2026arxiv-world,schmid2026arxiv-genrecon,zadaianchuk2026reconstruction-eccv} 
acquire and retain useful information to
understand the geometry of a scene from partial views under a limited computation
budget? This question connects two problems across computer 
vision and robotics: \emph{active reconstruction}, which selects the next best view 
to acquire, and \emph{keyframe selection for long-horizon reconstruction}, 
which selects a subset of views to retain as more frames are collected. 
Although these problems concern different stages of 
reconstruction, both require assessing how much information an observation contributes 
beyond the evidence available.

Existing active reconstruction methods represent scenes using occupancy 
grids \citep{zhou2021ral-fuel,guedon2022neurips-scone,guedon2023cvpr-macarons}, 
neural radiance fields \citep{lee2022ral-uncertainty,pan2022eccv-activenerf,yan2023iccv-active,feng2024cvpr-naruto,zhou2023dynpoint}, or Gaussian splatting \citep{jiang2023eccv-fisherrf,xie2025rss-gauss,li2025ral-activesplat,chen2025cvpr-activegamer,jin2025ral-activegs,xue2026cvpr-uncertainty,jeong2026icra-informative,jun2026eccv-sa}.
These methods typically model uncertainty within the current reconstruction and 
use it to guide subsequent observations. However, uncertainty defined over observed or 
instantiated geometry, i.e., reconstruction space, can provide an incomplete account 
of the priors we might have about unobserved space, resulting in suboptimal view selection.
Long-horizon reconstruction using generative models refers to
incrementally building and updating a 3D scene with a generative
model over an extended sequence of observations.
This setting poses a complementary challenge: retaining useful observations 
while keeping memory use and
per-update computation bounded as the input sequence grows. 
Keyframe selection is a longstanding approach in
simultaneous localization and mapping (SLAM)
\citep{mur2015tro-orb, maggio2026rss-vggt}, but remains less explored
in generative reconstruction.
Approaches that jointly incorporate all frames through weighted multi-diffusion 
\citep{bar2023arxiv-multidiffusion,li2026arxiv-mv} become computationally expensive 
as the sequence grows. Stream3D \citep{zhou26arxiv-stream3d} addresses memory 
growth through \emph{Adaptive Evidential Memory} (AEM).
AEM uses cross-attention between image features and 3D query tokens
to assign each observation an evidence weight for each token,
measuring how strongly that observation supports the token.
We refer to these observation-to-token weights, organized over
a shared set of spatially aligned 3D tokens, as \emph{evidence space}.
This representation makes it possible to aggregate support from
different views at corresponding 3D locations.
AEM retains a fixed number of high-evidence frame indices per token,
bounding the retained evidence while still processing every incoming
frame. Selecting keyframes therefore remains an important opportunity
for improving efficiency.

These limitations motivate a shared question: Can a unified evidence-space formulation of 
scene representation, uncertainty, and information gain guide which observations 
to acquire and retain? We introduce \emph{Matisse}, a training-free framework that
represents scenes in the evidence space of a pretrained generative 3D model.
From this formulation, Matisse derives uncertainty and information
gain to guide both active view selection and long-horizon keyframe selection.

First, to quantify how well the acquired observations support
each part of the scene, we introduce \emph{Evidential Uncertainty}
(EU). Matisse formulates Adaptive Evidential 
Memory as a maximum a posteriori (MAP) estimation over 3D 
latent tokens and derives uncertainty from the posterior 
covariance. This connects evidence aggregation and uncertainty estimation 
within a shared, spatially aligned latent representation. Because these tokens 
represent both observed and generatively completed geometry, EU 
can characterize uncertainty beyond the currently observed surfaces.

Second, to identify views that provide information beyond the
current observations, we introduce \emph{Evidential Information Gain}
(EIG). EIG measures the expected reduction in uncertainty from
additional evidence.
Inspired by FisherRF's information-driven planning
\citep{jiang2023eccv-fisherrf}, Matisse models how additional 
observations can reduce uncertainty. Whereas FisherRF evaluates information 
gain through a local approximation over reconstruction-space model parameters, 
Matisse evaluates it in the evidence space incorporating generative scene hypotheses. 
This allows acquiring and retaining views, while accounting for evidence relevant 
to both observed and occluded regions.

Third, we develop efficient \emph{evidence-space view selection}.
By defining EU and EIG in evidence space, we score candidate views without
expensive mesh generation. Batched GPU-accelerated
evaluation further reduces the amortized scoring time to 1.26 ms per candidate.

Fourth, we integrate these components into a scalable reconstruction system. 
Visibility-aware, object-balanced aggregation extends the information-gain 
criterion to multi-object scenes, accounting for occlusion while balancing 
the contributions of individual objects. An occupancy-stability 
termination test determines when to stop acquiring observations.

We evaluate Matisse on GSO30 single-object reconstruction \citep{Francis2022icra-gso}, YCB-V tabletop 
scenes \citep{Calli15-YCBobject}, and Replica room-scale environments \citep{Straub19arxiv-replica}. 
Matisse reduces the Chamfer 
distance (CD) versus the best prior active-reconstruction baselines by
$12.7\%$, $3.8\%$, and $9.2\%$, respectively. On GSO30, it reduces P-FID
by 9.6\% and runs 1.50$\times$ faster than the strongest prior active-view 
planner using the same Stream3D backend. Beyond active reconstruction, the 
same EIG criterion enables long-horizon keyframe selection. In the GSO30
keyframe selection experiment for long-horizon reconstruction, Matisse
achieves an $18.3\%$ lower CD than Stream3D and
approaches convergence using roughly $20\%$ as many views.
These results establish evidence-space reasoning
as a shared foundation for deciding where to acquire and which observations
to retain.


\section{Related Work}
\label{sec:related_work}

\myParagraph{Scene Representations for Active Reconstruction}
Scene representations have played a central
role in active reconstruction. Active reconstruction methods use geometric 
landmarks and occupancy maps
\citep{sim2005icra-global,bourgault2002iros-information,stachniss2005rss-information,
carlone2010iros-application,jadidi2015iros-mutual,bircher2016icra-receding,
zhou2021ral-fuel,guedon2022neurips-scone,guedon2023cvpr-macarons},
neural radiance fields
\citep{lee2022ral-uncertainty,pan2022eccv-activenerf,
yan2023iccv-active,feng2024cvpr-naruto},
or Gaussian splatting
\citep{jiang2023eccv-fisherrf,xie2025rss-gauss,
li2025ral-activesplat,chen2025cvpr-activegamer,
jin2025ral-activegs,xue2026cvpr-uncertainty,
jeong2026icra-informative,jun2026eccv-sa}
to represent scenes and guide view selection.
To reason beyond observed surfaces, learned completion methods predict
unobserved geometry
\citep{guedon2022neurips-scone,guedon2023cvpr-macarons,li2026cvpr-magician}.
MAGICIAN \citep{li2026cvpr-magician}, for example, predicts unseen
occupancy and converts it into Gaussian primitives for uncertainty rendering.
In contrast, we use a generative 3D foundation
model as the reconstruction backbone and maintain the scene as an implicit 3D evidence memory.
This formulation provides a learned geometric completion prior for occluded and unobserved regions
while permitting efficient uncertainty propagation as new observations arrive.

\myParagraph{Uncertainty and Information Gain}
Active perception requires an uncertainty measure
predicting the utility of candidate observations.
Existing methods guide acquisition using uncertainty or confidence measures
derived from predictive variance, rendering coverage, transmittance,
or geometric support
\citep{pan2022eccv-activenerf,chen2025cvpr-activegamer,
li2025ral-activesplat,jin2025ral-activegs}.
However, observing the most uncertain region does 
not necessarily yield the greatest information gain.
Information-driven approaches instead evaluate expected uncertainty
reduction
\citep{bourgault2002iros-information,stachniss2005rss-information,
carlone2010iros-application,jadidi2015iros-mutual}.
FisherRF \citep{jiang2023eccv-fisherrf} approximates information gain
over reconstruction-model parameters using the curvature of the rendering
likelihood, while GauSS-MI \citep{xie2025rss-gauss} evaluates mutual
information between Gaussian reliability and future observations.
GAVIS \citep{xue2026cvpr-uncertainty} combines anisotropic visibility 
fields with uncertainty-aware 
rendering to perform information-gain-based active mapping.
However, these methods formulate uncertainty and information gain 
with respect to the current scene representation or its rendered 
predictions. In contrast, Matisse derives uncertainty and information 
gain in the latent evidence space of a pretrained generative 3D model, 
enabling view selection over both observed and generatively completed 
geometry. Our formulation is
therefore global in scene-hypothesis space, rather than being restricted 
to a local geometry around the current model parameters.

\section{Problem Setup}
We study how a 3D reconstruction system can acquire and retain useful
information to recover scene geometry from partial views under a limited
computation budget. This problem involves two coupled decisions:
where to observe next and which observations to retain.
At each time step $t$, a mobile agent acquires an RGB-D observation
$(I_t, D_t)$ from camera view $v_t$ with camera pose $c_t \in \SEthree$.
To acquire useful information, the agent selects the next feasible
viewpoint $c_{t+1}$ based on its current estimate of the scene geometry.
To retain useful information, the system selects keyframes from the
observations collected so far for reconstruction.

\section{Preliminary}
\label{sec:preliminary}

\myParagraph{3D Generative Model}
Recent 3D generative models 
\citep{xiang2025cvpr-structured,chen26cvpr-sam,xiang26cvpr-native}
use a two-stage pipeline: a sparse-structure (SS) stage predicts coarse
occupancy, followed by a structured-latent (SLAT) stage that models geometry
and appearance. Both stages represent the object using spatially aligned
3D tokens in a canonical coordinate frame. Let $q\in\{1,\ldots,Q\}$ index
the canonical tokens; for example, $Q=16^3=4096$ in SAM3D.

\myParagraph{Adaptive Evidential Memory}
Our method builds on Adaptive Evidential Memory (AEM), introduced by
Stream3D \citep{zhou26arxiv-stream3d}, to combine observations at shared
spatial locations in the canonical 3D token grid.
For each acquired view $v \in \mathcal{V}_t$, an evidence weight
$M_v[q] \geq 0$ measures how
strongly the image features support the token at location $q$.
A low weight indicates weak support from that image, as may occur when
the corresponding region is occluded or outside the field of view.

For each token, AEM retains up to the $D$ highest evidence weights over all observed
views together with the indices of the views that supplied them. 
These paired entries form the evidence memory
$M \in \mathbb{R}^{Q \times D}$ and frame-index memory
$F \in \mathbb{N}^{Q \times D}$.
To select views for reconstruction, tokens vote for views recorded in
their frame-index memory, and these votes are aggregated across tokens to choose
$K$ views.
The resulting set $\mathcal{V}_t^* \subseteq \mathcal{V}_t$ supplies
the features used for reconstruction.

We focus on the sparse-structure stage, which predicts the object's
coarse occupancy \citep{chen26cvpr-sam}.
Let $z_\tau$ denote the noisy latent at denoising time $\tau$, distinct
from observation step $t$, and $\theta$ the pretrained generator parameters.
At each token $q$, the selected views contribute features
$V_\theta(z_\tau,v)[q]$, which are combined using their normalized
evidence weights:
\begin{equation}
    \bar{V}_\theta(z_\tau)[q]
    =
    \sum_{v\in\mathcal{V}^*_t}\bar{M}_v[q]V_\theta(z_\tau,v)[q],
    \qquad
    \bar{M}_v[q]
    =
    \frac{M_v[q]}{\sum_{u\in\mathcal{V}^*_t}M_u[q]}.
    \label{eq:stream3d-evidence-fusion}
\end{equation}
Thus, views with larger evidence weights contribute more strongly to 
the fused feature at each 3D location. 
The sparse-structure generator denoises the fused representation and 
decodes it into an occupancy prediction.


\section{Method}
\label{sec:method}
We propose an evidence-space view selection framework for 3D object
reconstruction that builds on Stream3D's canonical 
evidence memory \citep{zhou26arxiv-stream3d} over sparse 3D tokens. 
We derive evidential uncertainty from this memory
(Section~\ref{sec:method-uncertainty}) and use it to formulate information gain
for evaluating candidate views (Section~\ref{sec:method-information-gain}).
We then integrate these quantities into an efficient view-selection system
(Section~\ref{sec:method-system}).

\subsection{Evidential Uncertainty}
\label{sec:method-uncertainty}
Our key idea is to interpret Stream3D's association weights
probabilistically: a view that provides stronger support for a
3D token should contribute a more precise estimate of its
representation.
To formalize this intuition, let $x[q]\in\mathbb{R}^d$ denote the unknown, view-independent representation of
canonical token $q$.  We interpret Stream3D's association
weight $M_v[q]$ as indicating
how strongly we should trust view $v$'s observation of token $q$.
At fixed $\tau$, consider the Gaussian model
\begin{equation}
    p(x[q])=\mathcal{N}\!\left(\mu_0,\frac{\sigma^2}{\alpha}I\right),
    \qquad
    p\!\left(V_\theta(z_\tau,v)[q]\mid x[q]\right)
    =\mathcal{N}\!\left(x[q],\frac{\sigma^2}{M_v[q]}I\right),
    \label{eq:evidence-probabilistic-model}
\end{equation}
where $\mu_0\in\mathbb{R}^d$ is the prior mean, $\sigma^2>0$ is a shared
variance scale, $I$ is the $d\times d$ identity matrix, and $\alpha>0$ is
prior pseudo-evidence. A view with $M_v[q]=0$ contributes no likelihood term.
This interpretation allows us to derive uncertainty directly
from Stream3D's existing evidence memory. Assuming conditional independence
of the retained view features given $x[q]$, Gaussian conjugacy yields the
following result. Appendix~\ref{app:repeated-views} discusses the limitation of
the independence assumption.

\begin{greytheorem}[label=thm:map-estimate]{MAP Estimate}
Define the accumulated evidence score of token $q$ after observing
$\mathcal{V}^*_t$ as
\begin{equation}
    E_t[q]:=\sum_{v\in\mathcal{V}^*_t}M_v[q],
    \label{eq:accumulated-evidence-score}
\end{equation}
where larger values represent stronger or repeated observational support.
Given the prior and measurements with likelihoods as in 
Equation~\eqref{eq:evidence-probabilistic-model}, the resulting token posterior
is
\begin{equation}
\begin{aligned}
    p(x[q]\mid\mathcal{V}^*_t)
    &=\mathcal{N}(\mu_t[q],\Sigma_t[q]), \\
    \mu_t[q]
    &=\frac{\alpha\mu_0+\sum_{v\in\mathcal{V}^*_t}
             M_v[q]V_\theta(z_\tau,v)[q]}
            {\alpha+E_t[q]},
    &\Sigma_t[q]&=\frac{\sigma^2}{\alpha+E_t[q]}I.
\end{aligned}
    \label{eq:evidence-posterior}
\end{equation}
Its unique maximum-a-posteriori estimate is
$x^{\mathrm{MAP}}[q]=\mu_t[q]$.
\end{greytheorem}
The proof is provided in Appendix~\ref{app:proof-map}.

\myParagraph{Connection to Stream3D Fusion}
Substituting $E_t[q]=\sum_{v\in\mathcal{V}^*_t}M_v[q]$ and
Equation~\eqref{eq:stream3d-evidence-fusion} into the posterior mean from
Theorem~\ref{thm:map-estimate} gives the exact relation
$\mu_t[q]=\frac{\alpha}{\alpha+E_t[q]}\mu_0
+\frac{E_t[q]}{\alpha+E_t[q]}\bar{V}_\theta(z_\tau)[q]$.
When the prior pseudo-evidence is small relative to the accumulated observation
evidence, $\alpha\ll E_t[q]$, this relation reduces to
$x^{\mathrm{MAP}}[q]=\mu_t[q]\approx\bar{V}_\theta(z_\tau)[q]$.
Thus, in the evidence-dominated regime, Stream3D's normalized feature fusion is
approximately the MAP estimate of the view-independent canonical token
representation in Theorem~\ref{thm:map-estimate}.

Beyond the fused representation, the posterior covariance quantifies
the remaining uncertainty in each token. Following the A-optimality
criterion \citep{sim2005icra-global}, we summarize this uncertainty 
by the covariance trace and normalize it by its prior value.
\begin{greydefinition}[label=def:evidential-uncertainty]{Evidential Uncertainty}
Normalizing the posterior variance by its value before any view is observed
defines the bounded token uncertainty
\begin{equation}
    U_t[q]
    :=\frac{\operatorname{tr}\Sigma_t[q]}
            {\operatorname{tr}\Sigma_0[q]}
    =\frac{\alpha}{\alpha+E_t[q]}.
    \label{eq:uncertainty}
\end{equation}
\end{greydefinition}

\subsection{Evidential Information Gain}
\label{sec:method-information-gain}

For view selection, we evaluate information gain over the predicted occupied canonical
tokens, including those representing generatively
completed geometry. All subsequent token sums and visibility sets are restricted to these
occupied tokens. For each such token, the Gaussian posterior
$p(x[q]\mid\mathcal{V}^*_t)=\mathcal{N}(\mu_t[q],\Sigma_t[q])$
allows us to quantify the information gained from a candidate view as
the expected reduction in posterior entropy, as formalized below.

\begin{greytheorem}[label=thm:evidential-information-gain]{Evidential Information Gain}
For a candidate camera $c$ that contributes evidence $M_c[q]$, define the
token-wise information gain as its expected reduction in posterior entropy,
\begin{equation}
    \mathrm{IG}_t(q,c)
    :=H(x[q]\mid\mathcal{V}^*_t)
      -H(x[q]\mid\mathcal{V}^*_t,c)
    =\frac{1}{2}\log\frac{|\Sigma_t[q]|}
                              {|\Sigma_{t+1}[q;c]|}
    =\frac{d}{2}\log\!\left(
        1+\frac{M_c[q]}{\alpha+E_t[q]}
      \right).
    \label{eq:token-information-gain}
\end{equation}

Let $\Omega(c)$ denote the set of canonical tokens visible from
candidate camera $c$. Assume conditionally independent token latents, 
small incremental candidate evidence, and the binary-visibility approximation
$M_c[q]\approx\mathbf{1}[q\in\Omega(c)]$.  Then maximizing the candidate
information gain is approximated by maximizing
\begin{equation}
    \widehat{\mathrm{IG}}_t(c)
    :=
    \sum_{q\in\Omega(c)} U_t[q].
    \label{eq:uncertainty-gain}
\end{equation}
\end{greytheorem}

The proof of Theorem~\ref{thm:evidential-information-gain} is provided in
Appendix~\ref{app:proof-information-gain}. Appendix~\ref{app:gain-ranking-fidelity} (Table~\ref{tab:gain-ranking-fidelity})
empirically evaluates how closely Equation~\eqref{eq:uncertainty-gain} preserves
the logarithmic gain's candidate-view ranking under shared binary visibility. 
We compute $\Omega(c)$ by
projecting the canonical token grid into candidate camera $c$ and applying
a depth-buffer visibility test.

\subsection{Evidence-Space View Selection}
\label{sec:method-system}

Building on the evidential uncertainty and information gain derived above, 
we present a view-selection system that combines 
multi-object aggregation, efficient candidate scoring, 
and geometric exploration, and stops acquiring views 
when the reconstruction stabilizes.

\myParagraph{Multi-Object Formulation}
For a scene containing $N_{\mathrm{obj}}$ reconstructed objects, let
$\mathcal{Q}_t^{(i)}$ denote the occupied canonical tokens of object $i$.
We normalize each object's visible information gain by its token count,
$\widehat{\mathrm{IG}}_t^{(i)}(c)=
\sum_{q\in\Omega^{(i)}(c)}U_t^{(i)}[q]/|\mathcal{Q}_t^{(i)}|$,
and average across objects to obtain the scene-level score,
$\widehat{\mathrm{IG}}_t^{\mathrm{scene}}(c)=
\sum_{i=1}^{N_{\mathrm{obj}}}\widehat{\mathrm{IG}}_t^{(i)}(c)/N_{\mathrm{obj}}$.
This gives each object equal weight, preventing denser token grids from
dominating view selection. The score requires no additional prediction head:
it is computed directly from the evidence memory and decreases as evidence
accumulates over a fixed set of visible tokens.

\myParagraph{Efficient Evidence-Space View Selection}
View selection requires only sparse-structure decoding and online uncertainty updates
using Equation~\eqref{eq:uncertainty}, avoiding the expensive SLAT stage.
We further batch information-gain calculations across candidate views
on the GPU, achieving an amortized scoring time of $1.26$~ms per candidate.

\myParagraph{Occupancy-Stability Termination}
We stop acquiring views once the decoded sparse-structure occupancy stabilizes.
Let $\mathcal{P}_t$ and $\mathcal{P}_{t-1}$ be the occupied positions in the
canonical $64^3$ sparse-structure decoded voxel grid at two consecutive reconstruction steps.
For two sets $A$ and $B$, the symmetric difference
$A\mathbin{\triangle}B:=(A\setminus B)\cup(B\setminus A)$ contains elements
belonging to exactly one of the two sets. We measure the relative change as
$\Delta_t=|\mathcal{P}_t\mathbin{\triangle}\mathcal{P}_{t-1}|/|\mathcal{P}_t\cup\mathcal{P}_{t-1}|$
and terminate if $\Delta_t<\tau_{\mathrm{occ}}$.
We also impose a maximum view budget.  The test is applied only when a previous
reconstruction exists.  For a multi-object scene, it is evaluated independently
for each object: converged objects are frozen while active objects continue
along the shared camera trajectory.

\myParagraph{Exploration Factor}
Our approximate information-gain score does not explicitly account 
for proximity to previously acquired camera poses.
To encourage exploration, we weight information gain by viewpoint diversity.
Let $\mathcal{S}_t$ denote the camera poses acquired by step $t$ and
$\rho(c,c')$ the distance between two poses. The distance from candidate
$c$ to its nearest acquired pose is
$\rho(c,\mathcal{S}_t)=\min_{c'\in\mathcal{S}_t}\rho(c,c')$.
We define the exploration factor as
$w(c;\mathcal{S}_t)=1-\lambda\exp\!\bigl(-\rho(c,\mathcal{S}_t)^2/(2\ell^2)\bigr)$,
where $\lambda$ controls the maximum redundancy penalty and $\ell$ its
distance scale. This factor downweights views near previously acquired
poses and approaches one for more distant candidates.
Given a user-defined set of feasible camera poses $\mathcal{C}_t$
(see Appendix~\ref{app:implementation-settings} for our setup), we select
\begin{equation}
    \hat{c}_{t+1}
    =\arg\max_{c\in\mathcal{C}_t}s_t(c),
    \qquad
    s_t(c):=\widehat{\mathrm{IG}}_t^{\mathrm{scene}}(c)
    w(c;\mathcal{S}_t).
    \label{eq:onestep-score}
\end{equation}
Beyond this one-step policy, Matisse also supports receding-horizon planning
to account for future rewards, as detailed in
Appendix~\ref{app:receding-horizon-planning}.

\myParagraph{Keyframe Selection}
For keyframe selection, we apply the same view-selection rule to a given
sequence of posed observations, using the camera poses of unselected frames
as the candidate set $\mathcal{C}_t$.
Starting from an initial frame, we select the frame with the highest score
in Equation~\eqref{eq:onestep-score} and incorporate its observation into the
reconstruction to update the evidence memory and uncertainty.
We repeat this process with the remaining frames until the occupancy
stabilizes or the view budget is reached, retaining the selected frames
as keyframes.


\section{Experiments}
\label{sec:experiments}

\setlength{\textfloatsep}{\baselineskip}
\setlength{\dbltextfloatsep}{\baselineskip}
\setlength{\floatsep}{\baselineskip}
\setlength{\dblfloatsep}{\baselineskip}
\setlength{\intextsep}{\baselineskip}

We evaluate \emph{Matisse} on geometry and appearance reconstruction in
sparse-view 3D reconstruction. Sections~\ref{sec:exp-single-object} 
and~\ref{sec:exp-multi-object}
present single-object and multi-object reconstruction results,
respectively. Section~\ref{sec:exp-ablation} presents ablation
studies and analyses of evidential uncertainty.
\begin{figure}[!t]
    \centering
    \begin{minipage}{\linewidth}
    \centering
    {\footnotesize
    \setlength{\tabcolsep}{0pt}
    \begin{tabular}{*{7}{p{0.142857\linewidth}}}
        \centering GT & \centering Random & \centering FisherRF &
        \centering GauSS-MI & \centering GAVIS & \centering MAGICIAN &
        \centering Matisse-G\tabularnewline
    \end{tabular}\par}
    \includegraphics[width=\linewidth]{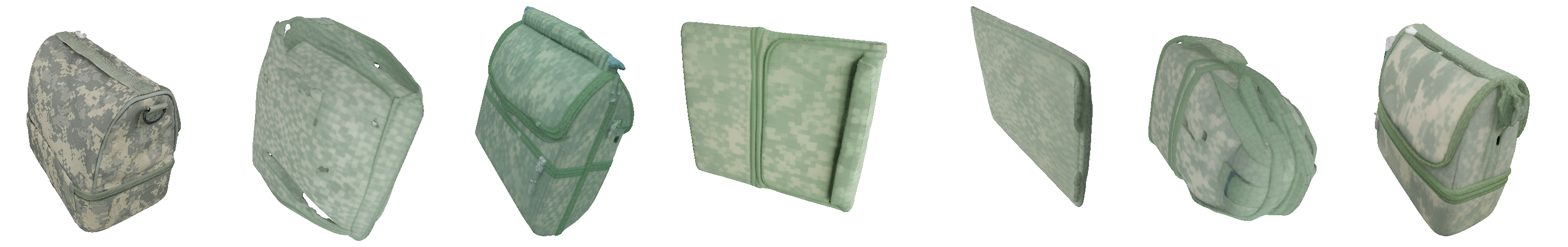}\par
    \includegraphics[width=\linewidth]{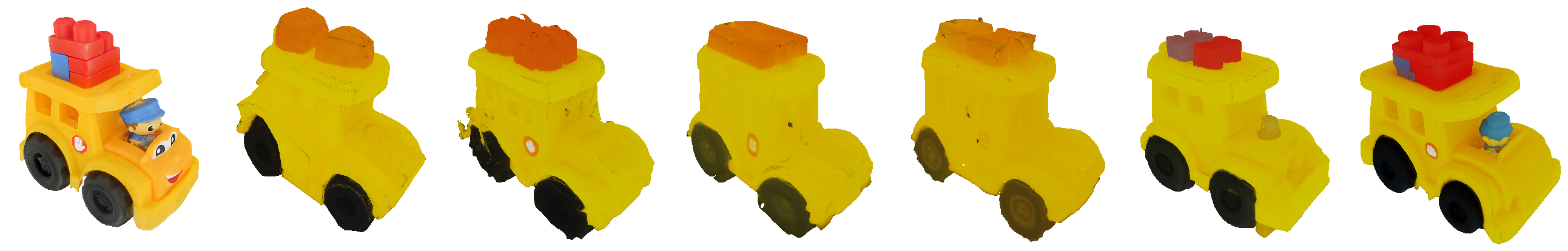}\par
    \includegraphics[width=\linewidth]{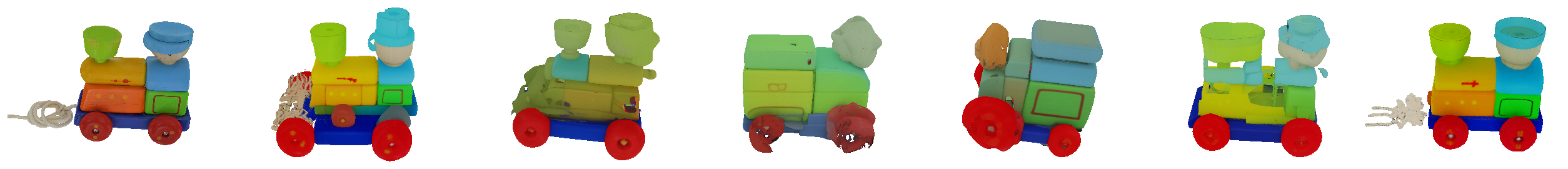}\par
    \includegraphics[width=\linewidth]{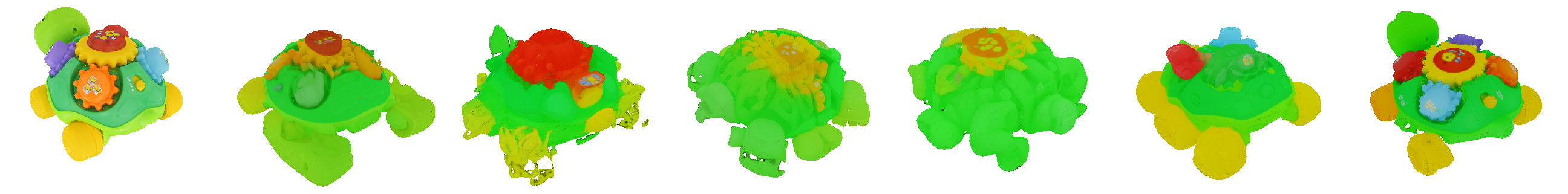}\par
    \end{minipage}
    \caption{\textbf{Qualitative comparisons on GSO30.} Matisse reconstructs geometry and appearance more
faithfully than the other active reconstruction baselines.}
    \label{fig:gso30-qualitative}
\end{figure}

\subsection{Experimental Setup}
\label{sec:exp-setup}

\myParagraph{Implementation}
Our experiments are run on two NVIDIA RTX 4090 GPUs. We use SAM3D as the
generative backbone, applying its four-step shortcut setting in both SS
and SLAT stages. We use Stream3D with its standard settings as the
multi-view fusion backbone. Each active run
starts with one view and makes four online updates,
each selecting the best next view and fusing the resulting observation
into the reconstruction. For YCB-V, we evaluate robustness to initialization
with five separate runs. Each uses a five-view budget and starts
from one of five initial views evenly spaced along the dataset trajectory.
We set $\alpha=1$; sensitivity to this choice is evaluated in
Appendix~\ref{app:alpha-sensitivity}. For receding-horizon baselines, we use a fixed
length $L=4$, discount $\gamma=0.9$, and occupancy-stability threshold
$\tau_{\mathrm{occ}}=0.1$. Dataset-specific camera and view-selection settings are provided in
Appendix~\ref{app:implementation-settings} (Table~\ref{tab:implementation}). Matisse-G
denotes Matisse with greedy mode and Matisse-RH denotes
Matisse with receding-horizon mode. Appendix~\ref{app:receding-horizon-planning} 
has more details on the receding-horizon planner.

\myParagraph{Datasets}
We use three complementary benchmarks. GSO30 \citep{kong2024eschernet-cvpr} is an evaluation subset of 30 single-object scenes
from Google Scanned Objects \citep{Francis2022icra-gso}.  YCB-V \citep{Calli15-YCBobject} contains 12 multi-object scenes, 
with 55 object instances from 21 categories. Following
\citet{Yu2022nips-MonoSDF,ni2024nips-phyrecon,ni2026iclr-g4splat}, we use
room and office scenes from Replica \citep{Straub19arxiv-replica},
manually selecting 54 objects from eight scenes to evaluate active
reconstruction in cluttered environments. All methods start from the
same initial observation within each trial.

\myParagraph{Baselines}
We compare Matisse with single-view feed-forward methods
SAM3D~\citep{chen26cvpr-sam} and TRELLIS.2~\citep{xiang26cvpr-native},
as well as TRELLIS.2+M.D., which does multi-diffusion~\citep{bar2023arxiv-multidiffusion}
without an evidence memory. For the single-view baselines SAM3D and TRELLIS.2, we use only
the initial image provided to the active methods.
Our active view selection baselines cover different combinations
of learned priors and information gain formulations:
GAVIS~\citep{xue2026cvpr-uncertainty} is a state-of-the-art method that uses neither,
FisherRF~\citep{jiang2023eccv-fisherrf} and
GauSS-MI~\citep{xie2025rss-gauss} use information gain without
a learned prior, and MAGICIAN~\citep{li2026cvpr-magician}
uses a learned prior without an information gain formulation.
We use the standard settings for each baseline, with the same 
3,500 iterations whenever 3DGS training is required.
We additionally include a random-walk planner and occupancy-grid planners
inspired by~\citet{bircher2016icra-receding}. Occupancy-grid planners use occupancy
predictions from the sparse-structure stage of SAM3D to serve as uniform visible-token coverage baselines. The greedy variant,
Occupancy-G, selects the next view that reveals the most
previously unseen occupancy. Occupancy-RH is the corresponding
receding-horizon variant.
For Replica, we disable MAGICIAN's beam search module 
due to its high computational cost under dense grid sampling (12.7 minutes per view selection).

\myParagraph{Metrics}
We evaluate geometry with Chamfer distance, mask IoU, and P-FID computed 
from PointNet++ features of reconstructed
point clouds following \citet{zhou26arxiv-stream3d}. GSO30 and YCB-V use bidirectional CD, 
whereas Replica uses one-sided CD because only partial ground truth meshes exist. 
We evaluate appearance with PSNR, SSIM and LPIPS following \citet{ni2026iclr-g4splat}. 
We also evaluate the end-to-end system runtime. 3DGS methods use their native open-source backends.

\subsection{Single-Object Reconstruction}
\label{sec:exp-single-object}

\begin{table*}[t]
\centering
\small
\setlength{\tabcolsep}{3.2pt}
\caption{GSO30 results. Best/second-best values are green/yellow.}
\label{tab:gso30}
\vspace{\baselineskip}
\sbox{\maintablebox}{%
\begin{tabular}{llccccccc}
\toprule
& & \multicolumn{3}{c}{Geometry} & \multicolumn{3}{c}{Appearance} & \multicolumn{1}{c}{Efficiency} \\
\cmidrule(lr){3-5}\cmidrule(lr){6-8}\cmidrule(lr){9-9}
Backend & Method & CD (mm) $\downarrow$ & IoU $\uparrow$ & P-FID $\downarrow$ & PSNR $\uparrow$ & SSIM $\uparrow$ & LPIPS $\downarrow$ & Runtime (s) $\downarrow$ \\
\midrule
\multirow{3}{*}{Feed-forward} & TRELLIS.2 \citep{xiang26cvpr-native} & 108.341 & 0.554 & 82.161 & 13.030 & 0.821 & 0.195 & 26.965{\scriptsize$\pm$14.716} \\
& TRELLIS.2 M.D. \citep{xiang26cvpr-native} & 115.192 & 0.571 & 110.274 & 13.095 & 0.829 & 0.192 & 62.283{\scriptsize$\pm$49.878} \\
& SAM3D \citep{chen26cvpr-sam} & 79.481 & 0.715 & 57.440 & 14.083 & \secondcell{0.848} & 0.173 & 3.444{\scriptsize$\pm$0.980} \\
\midrule
\multirow{4}{*}{3DGS} & FisherRF \citep{jiang2023eccv-fisherrf} & 76.267 & 0.583 & 99.486 & 12.031 & 0.765 & 0.244 & 49.121{\scriptsize$\pm$4.834} \\
& GauSS-MI \citep{xie2025rss-gauss} & 100.909 & 0.405 & 119.429 & 8.243 & 0.676 & 0.350 & 44.200{\scriptsize$\pm$0.627} \\
& GAVIS \citep{xue2026cvpr-uncertainty} & 92.276 & 0.528 & 108.910 & 10.710 & 0.754 & 0.263 & 100.032{\scriptsize$\pm$9.184} \\
& MAGICIAN \citep{li2026cvpr-magician} & 72.139 & 0.676 & 71.670 & 13.022 & 0.793 & 0.204 & 41.852{\scriptsize$\pm$0.781} \\
\midrule
\multirow{9}{*}{Stream3D} & Random & 62.078 & 0.736 & \secondcell{48.456} & 14.341 & 0.846 & 0.169 & 16.374{\scriptsize$\pm$1.566} \\
& Occupancy-G & 70.778 & 0.731 & 53.528 & 14.263 & \secondcell{0.848} & 0.168 & 16.929{\scriptsize$\pm$1.870} \\
& Occupancy-RH & 66.404 & 0.729 & 52.263 & 14.296 & 0.846 & 0.171 & 17.390{\scriptsize$\pm$1.953} \\
& FisherRF \citep{jiang2023eccv-fisherrf} & 65.866 & 0.716 & 53.443 & 13.941 & 0.843 & 0.176 & 55.566{\scriptsize$\pm$4.329} \\
& GauSS-MI \citep{xie2025rss-gauss} & 88.676 & 0.686 & 64.831 & 13.915 & 0.842 & 0.185 & 53.878{\scriptsize$\pm$2.785} \\
& GAVIS \citep{xue2026cvpr-uncertainty} & 87.598 & 0.687 & 64.394 & 13.894 & 0.844 & 0.187 & 95.679{\scriptsize$\pm$7.689} \\
& MAGICIAN \citep{li2026cvpr-magician} & 63.666 & 0.733 & 50.175 & \secondcell{14.392} & \secondcell{0.848} & \secondcell{0.166} & 23.965{\scriptsize$\pm$2.626} \\
\rowcolor{gray!20}& Matisse-G (Ours) & \bestcell{54.167} & \bestcell{0.765} & \bestcell{45.354} & \bestcell{14.542} & \bestcell{0.849} & \bestcell{0.159} & 16.023{\scriptsize$\pm$1.397} \\
\rowcolor{gray!20}& Matisse-RH (Ours) & \secondcell{56.228} & \secondcell{0.752} & 48.652 & 14.357 & 0.847 & 0.167 & 16.922{\scriptsize$\pm$1.714} \\
\bottomrule
\end{tabular}
}
\pgfmathparse{\the\linewidth/\the\wd\maintablebox}
\xdef\maintablescale{\pgfmathresult}
\scalebox{\maintablescale}{\usebox{\maintablebox}}
\end{table*}

\noindent
\begin{minipage}[t]{.63\textwidth}
\vspace{0pt}
\myParagraph{GSO30}
Table~\ref{tab:gso30} shows that Matisse-G leads all six reconstruction metrics,
reducing CD by $12.7\%$ versus the strongest baselines.
Unlike 3DGS-based methods, it completes unobserved geometry without costly optimization
before each selection step~\citep{jiang2023eccv-fisherrf,xie2025rss-gauss,xue2026cvpr-uncertainty},
making it $1.50\times$ faster than the fastest prior active-view planner.
MAGICIAN~\citep{li2026cvpr-magician} avoids online Gaussian optimization, but its beam
search repeatedly renders and scores future views, taking $3.96$ s per object.
Figure~\ref{fig:gso30-qualitative} shows that Matisse more faithfully reconstructs
geometry and appearance across four GSO30 objects.

\myParagraph{Keyframe Selection}
Figure~\ref{fig:gso30-view-convergence} compares keyframe-selection
methods on all 30 GSO30 objects using challenging casual trajectories,
each comprising 100 rendered views with partial visibility and intervals 
where the object is outside the camera's field of view. We consider three selection baselines: Random samples views
without replacement; Evenly-spaced uses uniformly spaced trajectory
indices; and Pose-diversity greedily maximizes the minimum angular
distance to previously selected views. Matisse selects up to eight views,
the three selection baselines select up to 48, and Stream3D processes
up to 100 views sequentially.
Comparing the fitted convergence plateaus, Matisse achieves $18.3\%$
lower CD than Stream3D and $21.9\%$ lower CD than the second-best baseline (Evenly-spaced).
It approaches convergence using roughly $20\%$ as many views
as Stream3D.
The corresponding YCB-V results are
provided in Appendix~\ref{app:keyframe-selection}.
\end{minipage}\hfill
\begin{minipage}[t]{.35\textwidth}
    \vspace{0pt}
    \centering
    \includegraphics[width=\linewidth]{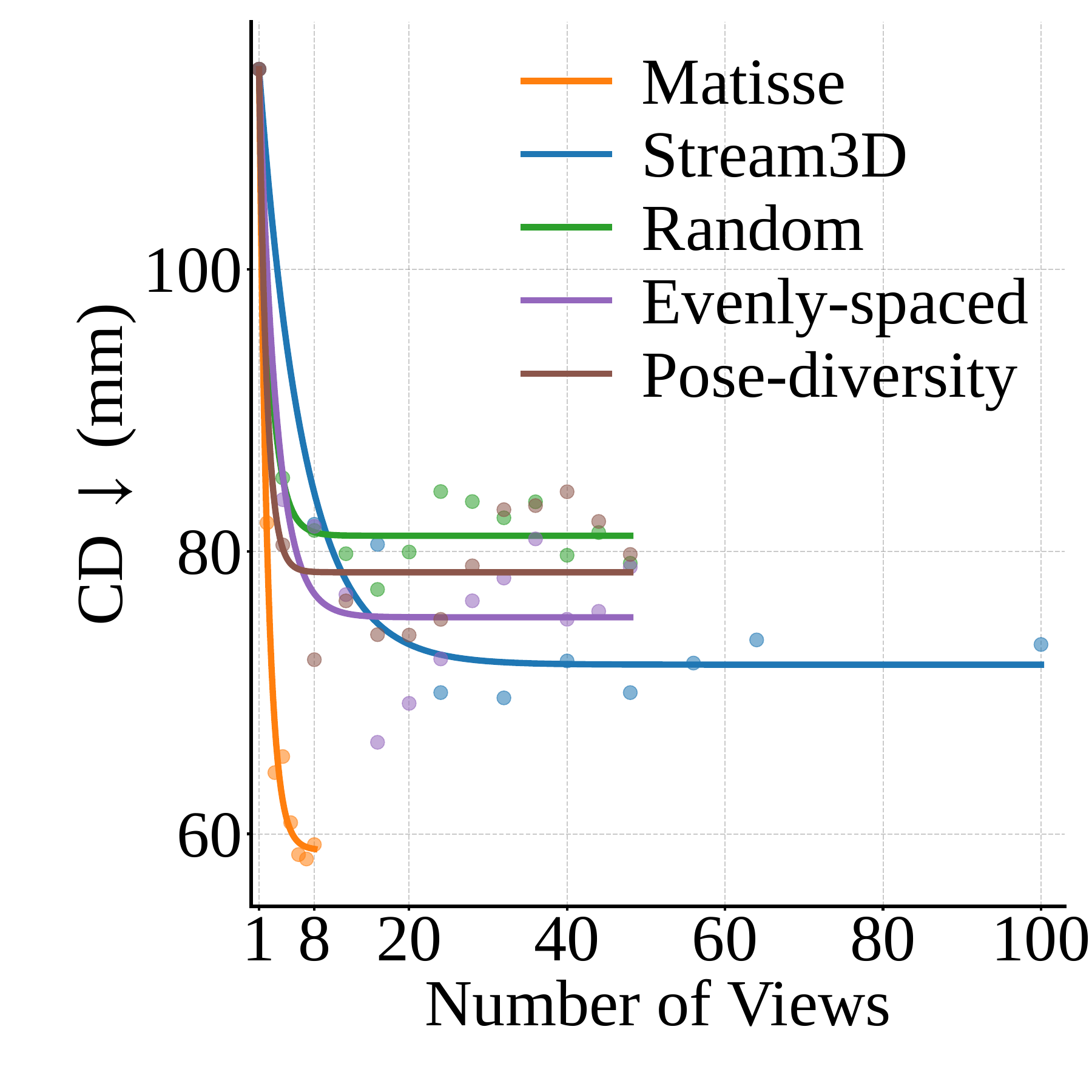}
    \captionof{figure}{GSO30 keyframe selection on challenging casual trajectories
    with exponential plateau fits.}
    \label{fig:gso30-view-convergence}
\end{minipage}
\par\medskip

\subsection{Multi-Object Reconstruction}
\label{sec:exp-multi-object}

\begin{table*}[t]
\centering
\small
\setlength{\tabcolsep}{2.8pt}
\caption{YCB-V results. Best/second-best values are green/yellow.}
\label{tab:ycbv}
\vspace{\baselineskip}
\scalebox{\maintablescale}{%
\begin{tabular}{llccccccc}
\toprule
& & \multicolumn{3}{c}{Geometry} & \multicolumn{3}{c}{Appearance} & \multicolumn{1}{c}{Efficiency} \\
\cmidrule(lr){3-5}\cmidrule(lr){6-8}\cmidrule(lr){9-9}
Backend & Method & CD (mm) $\downarrow$ & IoU $\uparrow$ & P-FID $\downarrow$ & PSNR $\uparrow$ & SSIM $\uparrow$ & LPIPS $\downarrow$ & Runtime (s) $\downarrow$ \\
\midrule
\multirow{3}{*}{Feed-forward} & TRELLIS.2 \citep{xiang26cvpr-native} & 16.768 & 0.665 & 66.609 & 18.658 & 0.915 & 0.087 & 99.969{\scriptsize$\pm$31.229} \\
& TRELLIS.2+M.D. \citep{xiang26cvpr-native} & 18.357 & 0.634 & 76.751 & 18.409 & 0.914 & 0.092 & 145.346{\scriptsize$\pm$52.187} \\
& SAM3D \citep{chen26cvpr-sam} & 7.043 & 0.869 & 22.561 & 20.421 & 0.922 & 0.065 & 14.513{\scriptsize$\pm$3.813} \\
\midrule
\multirow{4}{*}{3DGS} & FisherRF \citep{jiang2023eccv-fisherrf} & 18.494 & 0.332 & 137.698 & 16.145 & 0.915 & 0.104 & 109.827{\scriptsize$\pm$11.961} \\
& GauSS-MI \citep{xie2025rss-gauss} & 12.482 & 0.421 & 116.364 & 10.145 & 0.802 & 0.196 & 95.851{\scriptsize$\pm$9.754} \\
& GAVIS \citep{xue2026cvpr-uncertainty} & 12.281 & 0.681 & 106.955 & 14.585 & 0.884 & 0.114 & 143.018{\scriptsize$\pm$20.048} \\
& MAGICIAN \citep{li2026cvpr-magician} & 17.515 & 0.636 & 90.056 & 18.006 & 0.906 & 0.106 & 323.777{\scriptsize$\pm$66.196} \\
\midrule
\multirow{9}{*}{Stream3D} & Random & 6.705 & 0.877 & \secondcell{21.249} & 20.664 & 0.923 & 0.064 & 88.989{\scriptsize$\pm$19.424} \\
& Occupancy-G & 6.899 & 0.878 & 22.419 & 20.754 & 0.923 & 0.064 & 89.514{\scriptsize$\pm$20.168} \\
& Occupancy-RH & 6.822 & 0.880 & 21.731 & 20.835 & 0.923 & 0.063 & 89.816{\scriptsize$\pm$20.636} \\
& FisherRF \citep{jiang2023eccv-fisherrf} & 7.852 & 0.853 & 25.680 & 20.288 & 0.922 & 0.069 & 149.129{\scriptsize$\pm$20.894} \\
& GauSS-MI \citep{xie2025rss-gauss} & 6.731 & 0.877 & 22.039 & 20.614 & 0.922 & 0.064 & 139.062{\scriptsize$\pm$20.243} \\
& GAVIS \citep{xue2026cvpr-uncertainty} & 7.286 & 0.867 & 23.875 & 20.536 & 0.922 & 0.066 & 157.075{\scriptsize$\pm$18.592} \\
& MAGICIAN \citep{li2026cvpr-magician} & 7.835 & 0.853 & 25.421 & 20.241 & 0.921 & 0.068 & 92.614{\scriptsize$\pm$18.342} \\
\rowcolor{gray!20}& Matisse-G (Ours) & \bestcell{6.451} & \bestcell{0.885} & 21.450 & \secondcell{20.849} & \bestcell{0.923} & \bestcell{0.062} & 88.435{\scriptsize$\pm$18.773} \\
\rowcolor{gray!20}& Matisse-RH (Ours) & \secondcell{6.474} & \secondcell{0.885} & \bestcell{21.192} & \bestcell{20.912} & \secondcell{0.923} & \secondcell{0.063} & 91.185{\scriptsize$\pm$20.347} \\
\bottomrule
\end{tabular}
}
\end{table*}

\noindent\myParagraph{YCB-V}
Table~\ref{tab:ycbv} shows that Matisse remains effective
in multi-object scenes. Matisse-G achieves the best CD,
IoU, SSIM, and LPIPS, while Matisse-RH achieves the best
P-FID and PSNR. Both variants substantially outperform active view
selection baselines with 3DGS backends across all six
reconstruction metrics. Qualitative comparisons for two YCB-V multi-object 
scenes are provided in Appendix~\ref{app:ycbv-qualitative} (Figure~\ref{fig:ycbv-qualitative}).
Figure~\ref{fig:ycbv-view-convergence} in Appendix~\ref{app:view-budget-comparison} 
further compares view budgets on YCB-V. Across budgets ranging from 5 to 50 
frames, MAGICIAN's CD remains higher than that of Matisse using only five frames. 
This highlights a limitation of 3DGS-based reconstruction: adding more views 
cannot recover fully occluded surfaces, such as those in contact with the ground.

\myParagraph{Replica}
Table~\ref{tab:replica} in Appendix~\ref{app:replica-results}
further demonstrates Matisse's
effectiveness in cluttered indoor scenes.
Matisse-G achieves the best results across all six
reconstruction metrics. Compared with the strongest
baseline for each metric, it reduces CD and P-FID by
approximately $9.2\%$ and $5.2\%$, respectively, and
improves PSNR by $0.39$ dB. Qualitative comparisons on 
Replica are provided in Appendix~\ref{app:replica-qualitative}.

\myParagraph{Real-World Experiment}
We also deploy Matisse on a mobile robot, actively acquiring
19 frames to reconstruct 11 objects in a real scene
(Figure~\ref{fig:matisse-overview}(b)).
Hardware and acquisition details are provided in
Appendix~\ref{app:real-world-experiment}.

\begin{table*}[t]
\centering
\footnotesize
\setlength{\tabcolsep}{3.0pt}
\caption{YCB-V component ablations. Best/second-best values are green/yellow.}
\label{tab:ycbv-ablation}
\vspace{\baselineskip}
\scalebox{0.94}{\scalebox{\maintablescale}{%
\begin{tabular}{llcccccc}
\toprule
& & \multicolumn{3}{c}{Geometry} & \multicolumn{3}{c}{Appearance} \\
\cmidrule(lr){3-5}\cmidrule(lr){6-8}
& Method & CD (mm) $\downarrow$ & IoU $\uparrow$ & P-FID $\downarrow$ & PSNR $\uparrow$ & SSIM $\uparrow$ & LPIPS $\downarrow$ \\
\midrule
& w/o Information Gain & 7.403 & 0.866 & 23.633 & 20.489 & 0.921 & 0.066 \\
& w/o Exploration Factor & \secondcell{6.460} & 0.884 & \bestcell{20.938} & 20.831 & 0.923 & 0.063 \\
\rowcolor{gray!20}& Matisse-G (Ours) & \bestcell{6.451} & \bestcell{0.885} & 21.450 & \secondcell{20.849} & \bestcell{0.923} & \bestcell{0.062} \\
\rowcolor{gray!20}& Matisse-RH (Ours) & 6.474 & \secondcell{0.885} & \secondcell{21.192} & \bestcell{20.912} & \secondcell{0.923} & \secondcell{0.063} \\
\bottomrule
\end{tabular}
}}
\end{table*}

\Needspace{0.60\textheight}
\subsection{Ablations and Analysis}
\label{sec:exp-ablation}

\noindent
\begin{minipage}{\linewidth}
\begin{minipage}[t]{.58\textwidth}
\vspace{0pt}

\myParagraph{Contributions of Information Gain and Exploration}
We evaluate the contributions of information gain and exploration
on YCB-V (Table~\ref{tab:ycbv-ablation}).
Removing information gain leads to the largest overall degradation,
highlighting its importance for effective view selection.
The variant without the exploration factor remains competitive
and achieves the best P-FID. However, the full Matisse-G model
achieves the best CD, IoU, SSIM, and LPIPS, while Matisse-RH
achieves the best PSNR. These results support combining
information gain and exploration to improve reconstruction
quality.

\myParagraph{Sensitivity to Prior Pseudo-Evidence}
We find that $\alpha=1$ provides the best overall
performance, motivating its use in our experiments.
Detailed results are provided in
Appendix~\ref{app:alpha-sensitivity}.

\myParagraph{Analysis of Evidential Uncertainty}
We assess whether token-level EU reflects local geometric error using the GSO30 dataset.
We sample surface points from the predicted object meshes. 
Figure~\ref{fig:pointwise-eu-error} plots the EU of 
sampled surface points against their normalized geometric errors, measured as 
distances to the ground-truth mesh.
Across 899,341 valid surface samples from all 30 objects, normalized EU
and normalized geometric error have a Spearman rank correlation of
$\rho=0.266$, indicating a modest positive association.
Analyses of visibility, object symmetry, and model confidence under image
blur are provided in Appendix~\ref{app:uncertainty-analysis}.
\end{minipage}\hfill
\begin{minipage}[t]{.40\textwidth}
    \vspace{0pt}
    \centering
    \includegraphics[width=\linewidth]{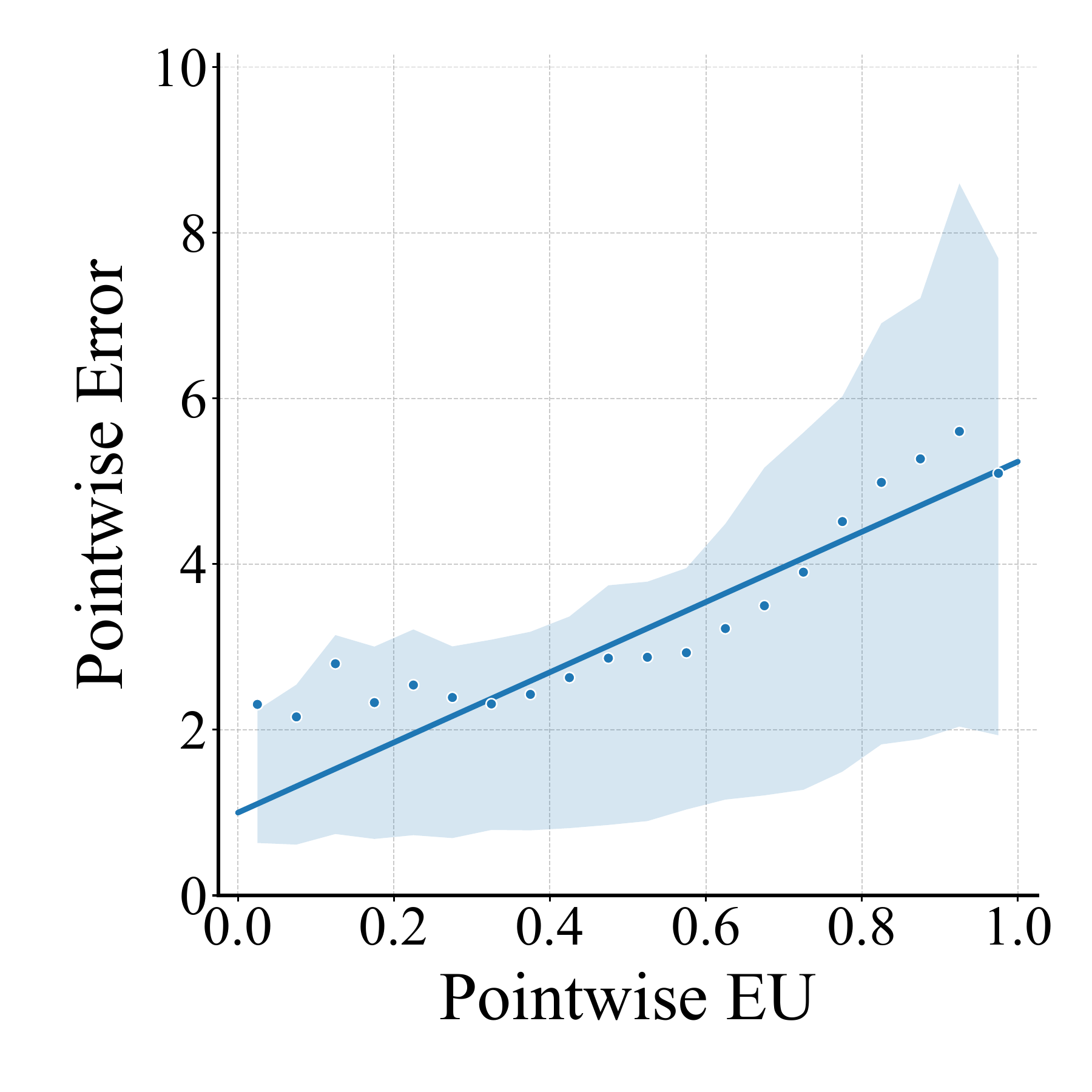}
    \captionof{figure}{\textbf{Pointwise EU versus geometric error on GSO30.}
    The x-axis is the normalized pointwise EU.
    The y-axis is the unsigned distance from a surface point to the
    nearest GT surface, expressed as a percentage of that object's GT
    bounding-box diagonal.}
    \label{fig:pointwise-eu-error}
\end{minipage}
\end{minipage}
\par\medskip

\section{Conclusion}
\label{sec:conclusion}

We presented Matisse, a training-free framework that unifies active 3D
reconstruction from sparse views and keyframe selection for long-horizon
reconstruction using a generative model. Matisse formulates Adaptive Evidential Memory as MAP
estimation, derives Evidential Uncertainty from the posterior covariance,
and uses Evidential Information Gain to guide view acquisition and retention.
Experiments on GSO30, YCB-V, and Replica demonstrate improved reconstruction
quality and computational efficiency across object-level and scene-level settings.
In the GSO30 keyframe selection experiment for long-horizon reconstruction,
Matisse achieves an $18.3\%$ lower CD than Stream3D
and approaches convergence using roughly $20\%$ as many views.

Together, these results support our central claim: evidence-space reasoning provides 
a shared foundation for active acquisition and keyframe selection by quantifying each view's 
information contribution conditioned on the current 3D memory.
Matisse thus offers a scalable path towards active 3D reconstruction.
Limitations and future directions are discussed in Appendices~\ref{app:limitations} and ~\ref{app:future_work}.

\clearpage

\bibliography{references/refs,references/myRefs}
\bibliographystyle{iclr2027_conference}

\clearpage
\appendix
\section{Proofs}
\label{app:proofs}

\subsection{Proof of Theorem~\ref{thm:map-estimate} (MAP Estimate)}
\label{app:proof-map}
By Bayes' rule and conditional independence of the observed view features, the
negative log-posterior, up to terms independent of $x[q]$, is
\begin{equation}
\begin{aligned}
    \mathcal{L}(x[q])
    &:=-\log p(x[q]\mid\mathcal{V}^*_t)+\mathrm{const.} \\
    &=\frac{\alpha}{2\sigma^2}\|x[q]-\mu_0\|_2^2
      +\sum_{v\in\mathcal{V}^*_t}
       \frac{M_v[q]}{2\sigma^2}
       \|V_\theta(z_\tau,v)[q]-x[q]\|_2^2, \\
    \nabla_{x[q]}\mathcal{L}
    &=\frac{1}{\sigma^2}\!\left(
       (\alpha+E_t[q])x[q]-\alpha\mu_0
       -\sum_{v\in\mathcal{V}^*_t}M_v[q]V_\theta(z_\tau,v)[q]\right).
\end{aligned}
\label{eq:map-objective}
\end{equation}
Setting the gradient to zero gives
$x^{\mathrm{MAP}}[q]=(\alpha\mu_0+
\sum_{v\in\mathcal{V}^*_t}M_v[q]V_\theta(z_\tau,v)[q])/
(\alpha+E_t[q])=\mu_t[q]$.  Moreover,
$\nabla^2\mathcal{L}=(\alpha+E_t[q])I/\sigma^2\succ0$ because $\alpha>0$;
hence this stationary point is the unique MAP solution.  The same expression
shows that posterior precision grows additively with Stream3D evidence.
\hfill$\square$

\subsection{Proof of Theorem~\ref{thm:evidential-information-gain} (Evidential Information Gain)}
\label{app:proof-information-gain}
For a $d$-dimensional Gaussian,
$H(x)=\tfrac12\log((2\pi e)^d|\Sigma|)$.  Assimilating candidate $c$ adds
$M_c[q]/\sigma^2$ to the precision, and therefore
$\Sigma_{t+1}[q;c]=\sigma^2I/(\alpha+E_t[q]+M_c[q])$.  Subtracting the two
Gaussian entropies proves Equation~\eqref{eq:token-information-gain}.  Assuming the
token latents are conditionally independent, entropy is additive and the view's
total information gain is the sum of its token-wise gains.  Define
$r_q=M_c[q]/(\alpha+E_t[q])$.  When each candidate supplies a small increment
relative to the current precision, Taylor expansion gives
\begin{equation}
\begin{aligned}
    \mathrm{IG}_t(c)
    &=\frac{d}{2}\sum_q\log(1+r_q) \\
    &=\frac{d}{2\alpha}\sum_q M_c[q]U_t[q]
      +\mathcal{O}\!\left(\sum_q r_q^2\right) \\
    &\approx\frac{d}{2\alpha}\widehat{\mathrm{IG}}_t(c),
\end{aligned}
    \label{eq:information-gain-approximation}
\end{equation}
where the last line replaces the unknown future association by binary geometric
visibility, $M_c[q]\approx\mathbf{1}[q\in\Omega(c)]$.  We compute
$\Omega(c)$ by projecting the canonical token grid into candidate camera $c$
and applying a depth-buffer visibility test.  Since $d/(2\alpha)>0$ is constant
across candidates, removing it does not change the maximizing view.  This proves
Equation~\eqref{eq:uncertainty-gain}.
Equation~\eqref{eq:uncertainty-gain} should therefore be read as a first-order,
visibility-based approximation to Bayesian information gain.  It avoids
predicting the content and confidence of an unobserved image while retaining
the desired diminishing return for tokens that are already well supported.
\hfill$\square$

\section{Receding-Horizon Planning}
\label{app:receding-horizon-planning}
To account for future rewards, we use receding-horizon planning
\citep{bircher2016icra-receding,schmid2020ral-efficient,li2026cvpr-magician}, at
each step searching feasible candidate sequences $b=(c_1,\ldots,c_L)$ of
horizon $L$. A branch is scored by its discounted cumulative gain,
\begin{equation}
    R_t(b)
    =
    \sum_{h=1}^{L}
    \gamma^{h-1}
    s_t^{(h)}(c_h),
    \label{eq:rh-score}
\end{equation}
where $\gamma\in(0,1]$ is the rollout discount.  At depth $h$, the exploration
factor includes the acquired cameras and the preceding branch cameras
$(c_1,\ldots,c_{h-1})$, while the uncertainty gain remains fixed to the current
evidence field.  We execute only the first view of the best branch,
\begin{equation}
    c_{t+1}
    =
    \operatorname*{first}
    \left[
    \arg\max_{b=(c_1,\ldots,c_L)} R_t(b)
    \right],
    \label{eq:rh-selection}
\end{equation}
then update Stream3D's evidence memory and replan.  Hypothetical views never
update the evidence field; only executed observations do so.

\section{Additional Experimental Details and Results}
\label{app:experimental-details}

\subsection{Quantitative Results on Replica}
\label{app:replica-results}

Table~\ref{tab:replica} reports reconstruction results on Replica.
Compared with the strongest baseline for each metric, Matisse-G reduces
CD and P-FID by $9.2\%$ and $5.2\%$, respectively, and improves PSNR
by $0.39$ dB. These gains indicate improved geometry and appearance
reconstruction in cluttered indoor scenes.

\begin{center}
\centering
\small
\setlength{\tabcolsep}{3.0pt}
\captionof{table}{Results on the scene-level Replica dataset.}
\label{tab:replica}
\vspace{\baselineskip}
\scalebox{\maintablescale}{%
\begin{tabular}{llcccccc}
\toprule
& & \multicolumn{3}{c}{Geometry} & \multicolumn{3}{c}{Appearance} \\
\cmidrule(lr){3-5}\cmidrule(lr){6-8}
Backend & Method & CD (mm) $\downarrow$ & IoU $\uparrow$ & P-FID $\downarrow$ & PSNR $\uparrow$ & SSIM $\uparrow$ & LPIPS $\downarrow$ \\
\midrule
\multirow{4}{*}{3DGS} & FisherRF \citep{jiang2023eccv-fisherrf} & 45.186 & 0.278 & 106.041 & 20.712 & 0.964 & 0.054 \\
& GauSS-MI \citep{xie2025rss-gauss} & 34.642 & 0.320 & 103.467 & 11.582 & 0.866 & 0.121 \\
& GAVIS \citep{xue2026cvpr-uncertainty} & 32.732 & 0.677 & 64.571 & 21.699 & 0.963 & 0.038 \\
& MAGICIAN \citep{li2026cvpr-magician} & 63.802 & 0.556 & 79.714 & 19.061 & 0.948 & 0.061 \\
\midrule
\multirow{6}{*}{Stream3D} & Random & 23.820 & 0.708 & 45.440 & 23.816 & 0.966 & 0.037 \\
& FisherRF \citep{jiang2023eccv-fisherrf} & 27.301 & 0.713 & 49.320 & 23.843 & 0.966 & 0.037 \\
& GauSS-MI \citep{xie2025rss-gauss} & \runnerupcell{21.028} & 0.738 & \runnerupcell{39.764} & 23.876 & 0.966 & 0.034 \\
& GAVIS \citep{xue2026cvpr-uncertainty} & 23.395 & 0.715 & 43.508 & 23.789 & 0.966 & 0.037 \\
& MAGICIAN \citep{li2026cvpr-magician} & 23.195 & \runnerupcell{0.746} & 40.243 & \runnerupcell{24.056} & \runnerupcell{0.968} & \runnerupcell{0.032} \\
\rowcolor{gray!20}& Matisse-G (Ours) & \bestcell{19.103} & \bestcell{0.761} & \bestcell{37.686} & \bestcell{24.448} & \bestcell{0.968} & \bestcell{0.031} \\
\bottomrule
\end{tabular}
}
\end{center}

\subsection{Dataset-Specific Implementation Settings}
\label{app:implementation-settings}

Table~\ref{tab:implementation} summarizes the settings for each dataset.
All three datasets use a five-view budget, while candidate sampling
and the distance metric follow the respective action spaces.

\begin{table}[htb]
\centering
\footnotesize
\setlength{\tabcolsep}{3pt}
\renewcommand{\arraystretch}{1.0}
\caption{Dataset-specific implementation settings.}
\label{tab:implementation}
\vspace{\baselineskip}
\begin{tabular}{@{}lccc@{}}
\toprule
Setting & GSO30 & YCB-V & Replica \\
\midrule
Action space & $1$ m cube & $30^\circ$ spherical cap & $6$ m cube \\
Resolution & $256\times256$ & $640\times480$ & $640\times480$ \\
Position sampling & Cartesian grid & Equal-area cap & Cartesian grid \\
Viewing direction & Toward object center & Toward scene center & Toward object center \\
Budget & 5 views per object & 5 views per scene & 5 views per object \\
Multi-object aggregation & No & Yes & No \\
Candidate poses & $3\times3\times3$ & $30$ & $8\times8\times8$ \\
Distance $\rho(c,c')$ & $\|\mathbf{p}_c-\mathbf{p}_{c'}\|_2$ & $\arccos(\mathbf{u}_c^\top\mathbf{u}_{c'})$ & $\|\mathbf{p}_c-\mathbf{p}_{c'}\|_2$ \\
Exploration penalty $\lambda$ & $0.5$ & $0.5$ & $0.5$ \\
Penalty scale $\ell$ & $0.5$ m & $30^\circ$ & $0.5$ m \\
\bottomrule
\end{tabular}
\par\smallskip
\begin{minipage}{0.88\linewidth}
\footnotesize
$\mathbf{p}_c$: camera center;
$\mathbf{u}_c=(\mathbf{p}_c-\mathbf{o})/\|\mathbf{p}_c-\mathbf{o}\|_2$,
with scene center $\mathbf{o}$. Distances use metres or degrees;
candidate counts precede validity filtering.
\end{minipage}
\end{table}

Each candidate pose consists of a sampled camera position and
an orientation directed toward the object or scene center in the aggregated depth pointmap.
For Cartesian sampling, camera positions lie on a regular
3D grid within the specified cube.
For YCB-V, we sample 30 camera positions over the spherical cap
using equal-area sampling.

\subsection{Real-World Experiment}
\label{app:real-world-experiment}

For the real-world demonstration in Figure~\ref{fig:matisse-overview}(b),
we deploy Matisse on an AgileX Tracer 2.0 mobile robot equipped with an
Intel RealSense Depth Camera D455. The system actively captures 19 frames to
reconstruct 11 objects. Because a single frame may provide evidence for
multiple objects, we score candidate views using the scene-level aggregation
described under \emph{Multi-Object Formulation} in
Section~\ref{sec:method-system}. We partition the objects using a
greedy, order-dependent procedure: the first pending object initializes a
group, and each subsequent object joins only if its center lies within $0.5$ m
of every object center already in the group. The resulting groups are
processed sequentially.

\subsection{Sensitivity to Prior Pseudo-Evidence}
\label{app:alpha-sensitivity}
We evaluate four logarithmically spaced values of $\alpha$
under the same GSO30 protocol as the main benchmark.
As shown in Table~\ref{tab:gso30-alpha-sensitivity},
performance varies modestly across the tested values,
with $\alpha=1$ achieving the best overall results.
We therefore use $\alpha=1$ as a practical default 
for all datasets.

\begin{table}[htbp]
\centering
\small
\setlength{\tabcolsep}{3.0pt}
\caption{Sensitivity analysis of prior pseudo-evidence $\alpha$ on GSO30. 
Green and yellow mark the best and second-best settings; the selected $\alpha=1$ row is
gray.}
\label{tab:gso30-alpha-sensitivity}
\vspace{\baselineskip}
\begin{tabular}{lcccccc}
\toprule
& \multicolumn{3}{c}{Geometry} & \multicolumn{3}{c}{Appearance} \\
\cmidrule(lr){2-4}\cmidrule(lr){5-7}
$\alpha$ & CD (mm) $\downarrow$ & IoU $\uparrow$ & P-FID $\downarrow$ & PSNR $\uparrow$ & SSIM $\uparrow$ & LPIPS $\downarrow$ \\
\midrule
0.01 & 58.855 & 0.7568 & 48.952 & 14.410 & \runnerupcell{0.8483} & 0.1630 \\
0.1 & 57.483 & 0.7561 & 49.078 & 14.379 & 0.8480 & 0.1633 \\
\rowcolor{gray!20}1.0 & \bestcell{54.167} & \bestcell{0.7650} & \bestcell{45.354} & \bestcell{14.542} & \bestcell{0.8491} & \bestcell{0.1593} \\
10.0 & \runnerupcell{54.673} & \runnerupcell{0.7593} & \runnerupcell{47.009} & \runnerupcell{14.541} & 0.8482 & \runnerupcell{0.1605} \\
\bottomrule
\end{tabular}
\end{table}

\subsection{View-Budget Comparison on YCB-V}
\label{app:view-budget-comparison}

\begin{figure}[htbp]
    \centering
    \includegraphics[width=0.40\linewidth]{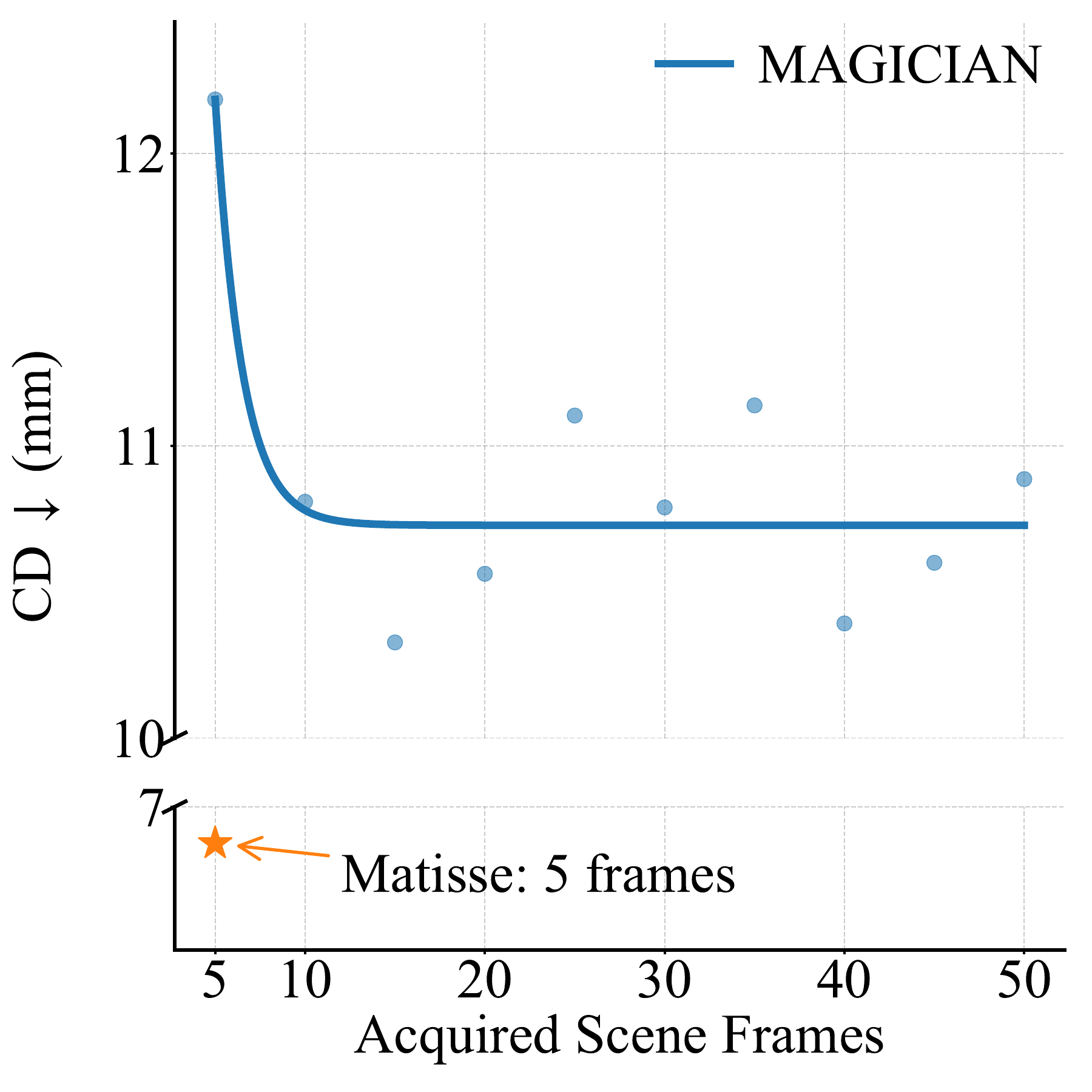}
    \caption{\textbf{View-budget comparison on YCB-V scene 48.}
    Blue points show MAGICIAN's measured Chamfer distance from
    5 to 50 acquired frames; the blue curve is a descriptive
    exponential decay fit with a plateau. The orange star marks
    Matisse's 5-frame result. The broken vertical axis separates
    the two CD ranges. Lower CD is better.}
    \label{fig:ycbv-view-convergence}
\end{figure}

Figure~\ref{fig:ycbv-view-convergence} compares MAGICIAN at budgets of
5--50 frames with Matisse using five frames on YCB-V scene 48.
MAGICIAN's CD decreases initially and then fluctuates around a plateau,
remaining higher than Matisse's five-frame result even at 50 frames.

\subsection{Additional Keyframe-Selection Results on YCB-V}
\label{app:keyframe-selection}

We evaluate keyframe selection on all 55 object instances across the
12 YCB-V scenes. For each scene, we generate a 100-view casual 
trajectory using rendered RGB images, ground-truth depth, and 
known camera poses. These trajectories include partial visibility 
and intervals when objects fall outside the camera's field of view.

\begin{figure}[htbp]
    \centering
    \includegraphics[width=0.48\linewidth]{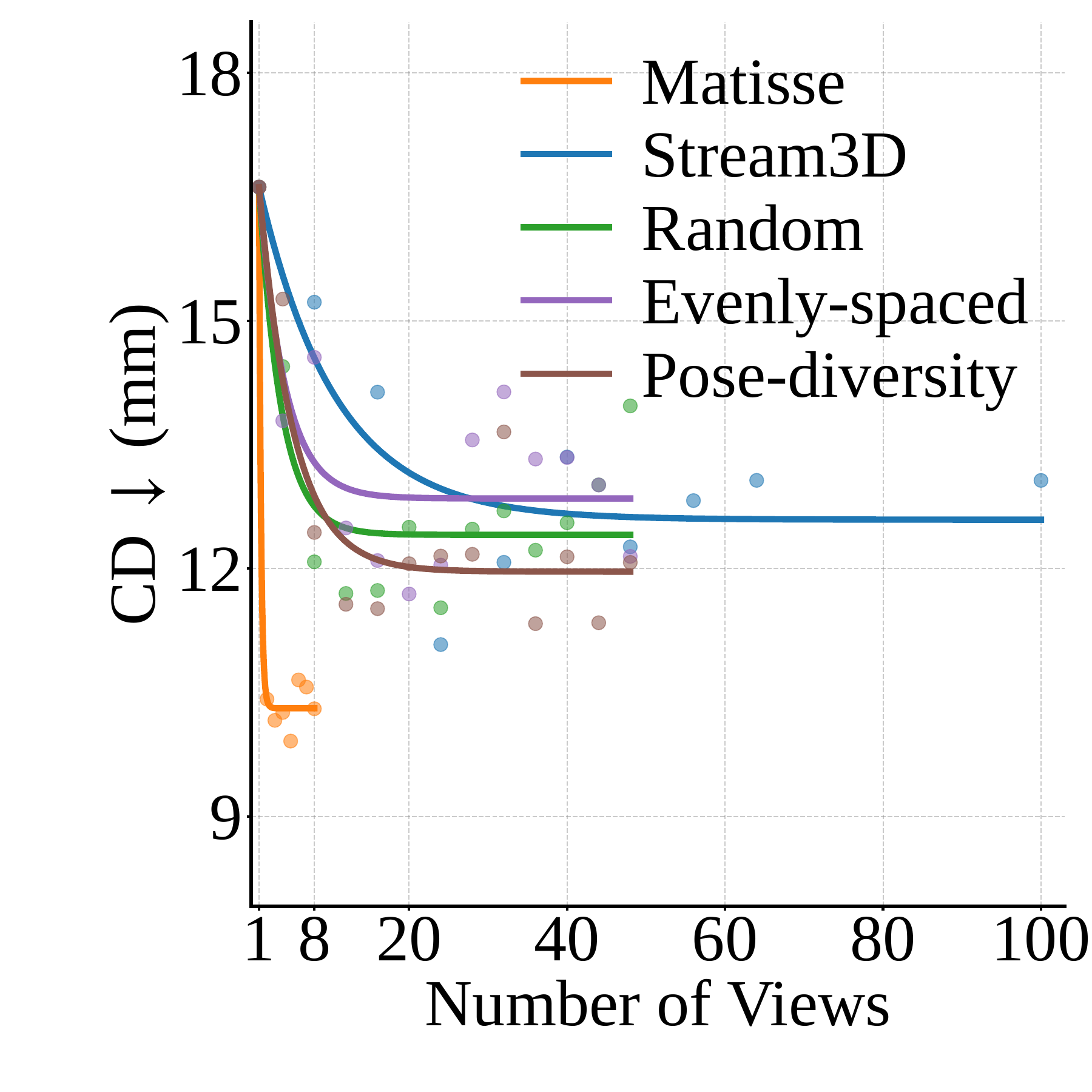}
    \caption{YCB-V keyframe selection on challenging casual trajectories
    with exponential plateau fits.}
    \label{fig:ycbv-casual-view-convergence}
\end{figure}

Figure~\ref{fig:ycbv-casual-view-convergence} reports mean CD, weighted
equally over all 55 instances. Matisse achieves $10.31$ mm with eight
selected views, compared with $13.07$ mm for Stream3D after processing
100 views sequentially. Random, Evenly-spaced, and Pose-diversity achieve
$12.08$--$13.97$ mm with 48 selected views. Matisse thus achieves lower
reconstruction error while using substantially fewer views.

\subsection{Qualitative Results on YCB-V}
\label{app:ycbv-qualitative}

\begin{figure}[htbp]
    \centering
    {\footnotesize
    \setlength{\tabcolsep}{0pt}
    \begin{tabular}{*{7}{p{0.142857\linewidth}}}
        \centering GT & \centering Random & \centering FisherRF &
        \centering GauSS-MI & \centering GAVIS & \centering MAGICIAN &
        \centering Matisse-G\tabularnewline
    \end{tabular}\par}
    \includegraphics[width=\linewidth]{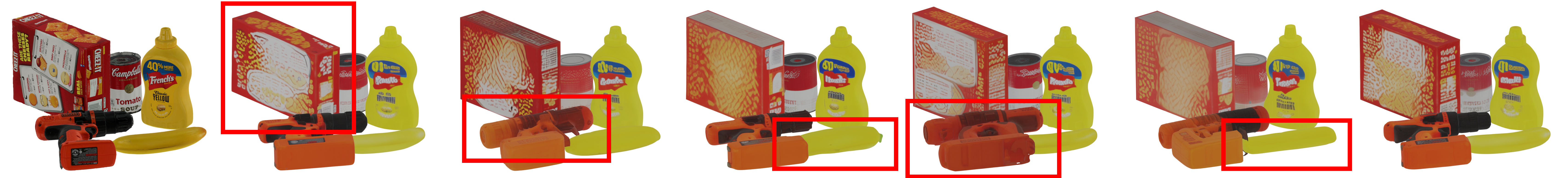}\par
    \includegraphics[width=\linewidth]{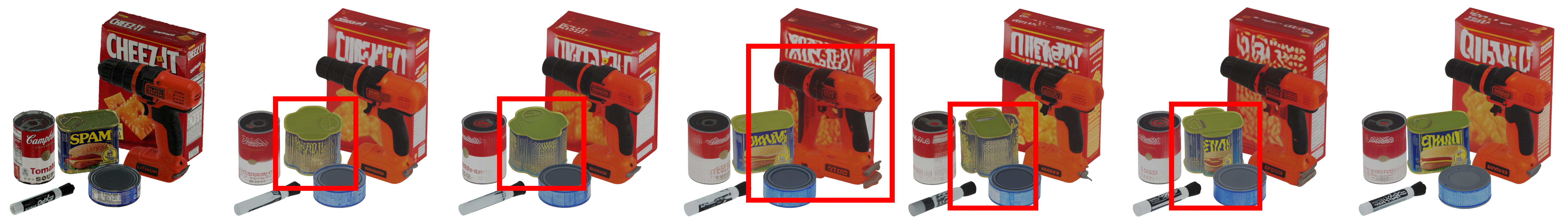}\par
    \caption{\textbf{Qualitative comparisons on YCB-V.}
    Rows show scene 50 from the fifth camera and scene 59 from the third
    camera. Columns show ground truth, Random, FisherRF, GauSS-MI,
    GAVIS, MAGICIAN, and Matisse-G; all reconstruction methods use the
    Stream3D backend. Predictions are vertex-colored meshes with registration
    and selected object-orientation corrections for visualization.
    Ground-truth geometry appears only in the GT column.}
    \label{fig:ycbv-qualitative}
\end{figure}

Figure~\ref{fig:ycbv-qualitative} compares reconstructed geometry and
appearance for two multi-object scenes against the ground truth.

\subsection{Qualitative Results on Replica}
\label{app:replica-qualitative}

\begin{figure}[htbp]
    \centering
    {\footnotesize
    \setlength{\tabcolsep}{0pt}
    \begin{tabular}{*{7}{p{0.142857\linewidth}}}
        \centering GT & \centering Random & \centering FisherRF &
        \centering GauSS-MI & \centering GAVIS & \centering MAGICIAN &
        \centering Matisse-G\tabularnewline
    \end{tabular}\par}
    \includegraphics[width=\linewidth]{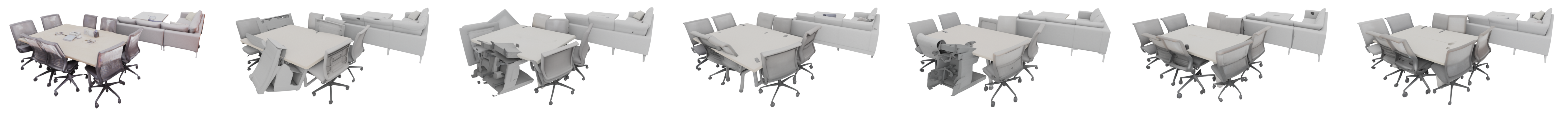}\par
    \includegraphics[width=\linewidth]{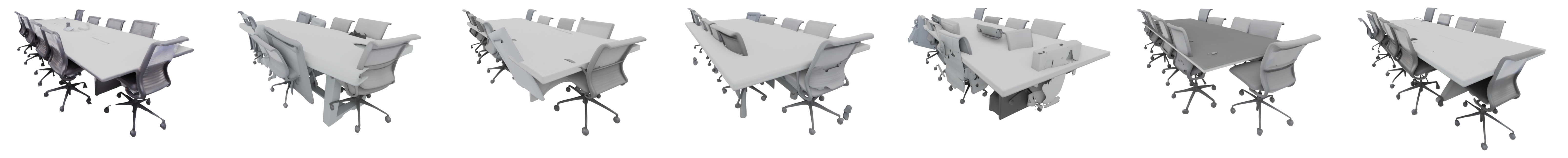}
    \caption{\textbf{Qualitative mesh comparisons on Replica.}
    Rows show office\_3 (top) and office\_4 (bottom).
    Columns show ground truth, Random, FisherRF, GauSS-MI, GAVIS,
    MAGICIAN, and Matisse-G. All reconstruction methods use the
    Stream3D backend. Registered meshes of the selected objects are
    shown from a common viewpoint within each row.}
    \label{fig:replica-qualitative}
\end{figure}

Figure~\ref{fig:replica-qualitative} compares reconstructed objects in
two Replica office scenes. The common viewpoints show differences in
the recovered furniture geometry relative to the ground truth.

\subsection{Approximation of Information Gain}
\label{app:gain-ranking-fidelity}
We compare the practical score in Equation~\eqref{eq:uncertainty-gain} with
the logarithmic information gain derived from Equation~\eqref{eq:token-information-gain},
using the Table~\ref{tab:gso30} protocol.
Table~\ref{tab:gain-ranking-fidelity} shows high agreement among the
top-ranked candidates, with 100\% top-1 agreement across the evaluated checkpoints.
These results support the logarithmic linearization conditional on binary
visibility.

\begin{center}
\begin{minipage}{\linewidth}
\centering
\small
\captionof{table}{\textbf{Analysis of information-gain approximation on GSO30.}
Top-$k$ agreement is the fraction of shared candidates in the two top-$k$
sets of linear and logarithmic scores.}
\label{tab:gain-ranking-fidelity}
\vspace{\baselineskip}
\begin{tabular}{lccc}
\toprule
Comparison & Top-1 agreement & Top-5 agreement & Top-10 agreement \\
\midrule
Linear vs. logarithmic gain & 100\% & 99.83\% & 99.50\% \\
\bottomrule
\end{tabular}
\end{minipage}
\end{center}

\section{Analysis of Predicted Uncertainty}
\label{app:uncertainty-analysis}

We analyze Matisse's predicted uncertainty from three complementary
perspectives: its relationship to visibility, its consistency with object
symmetry, and its relationship to model confidence as visual detail is lost.

\begin{figure}[htbp]
    \centering
    \begin{minipage}{0.32\textwidth}
        \centering
        \includegraphics[width=\linewidth]{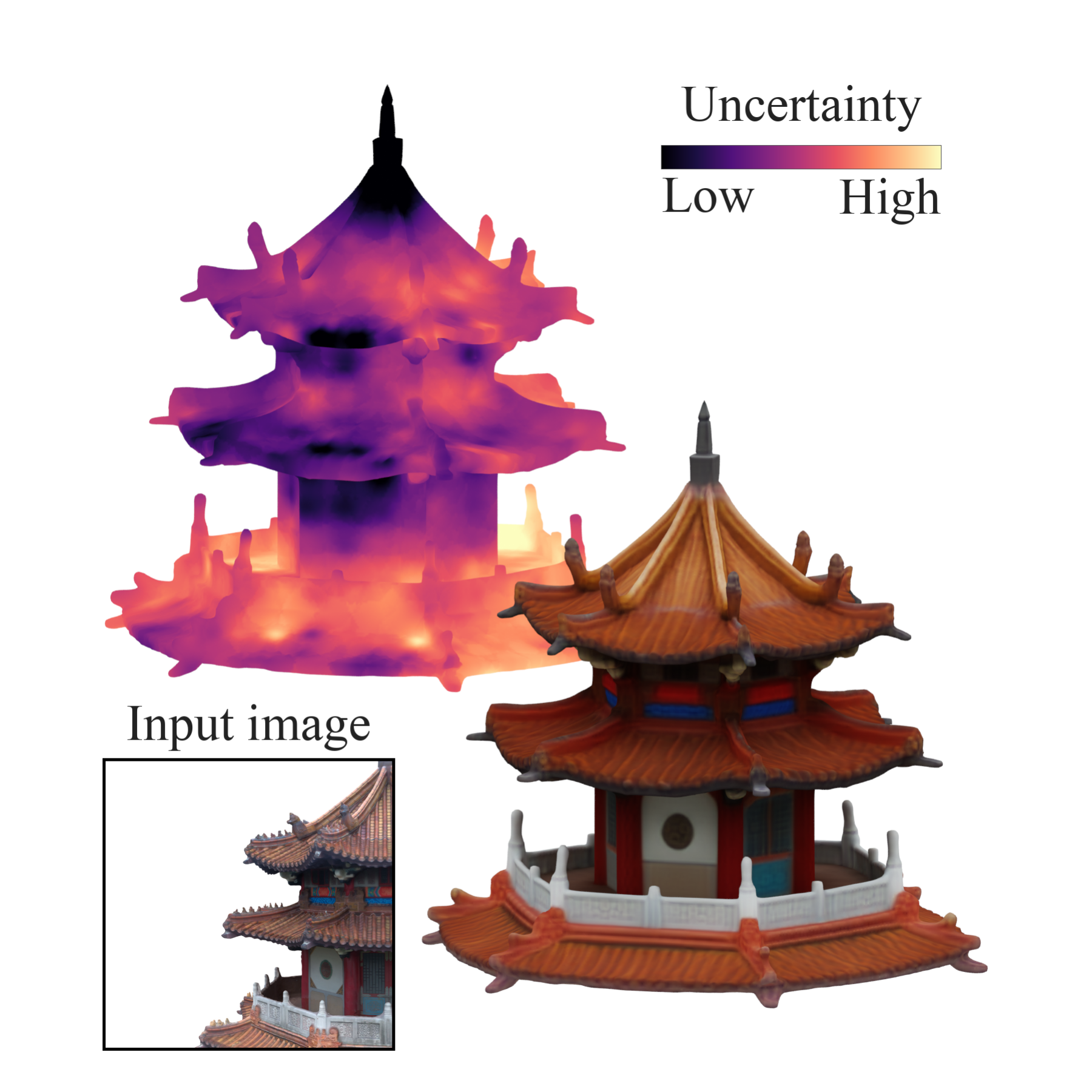}
        \textbf{(a)} Visibility
    \end{minipage}\hfill
    \begin{minipage}{0.32\textwidth}
        \centering
        \includegraphics[width=\linewidth]{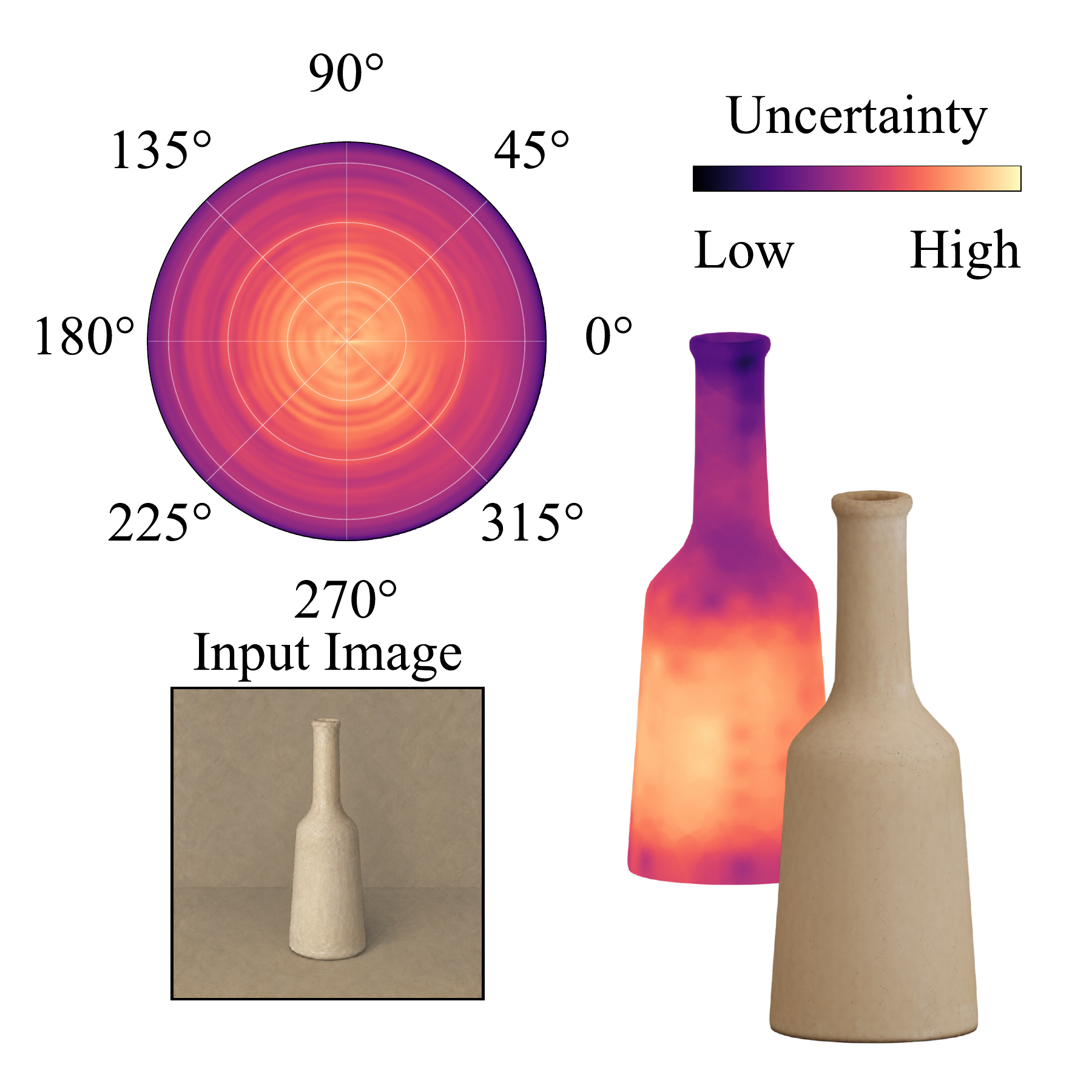}
        \textbf{(b)} Symmetry
    \end{minipage}\hfill
    \begin{minipage}{0.32\textwidth}
        \centering
        \includegraphics[width=\linewidth]{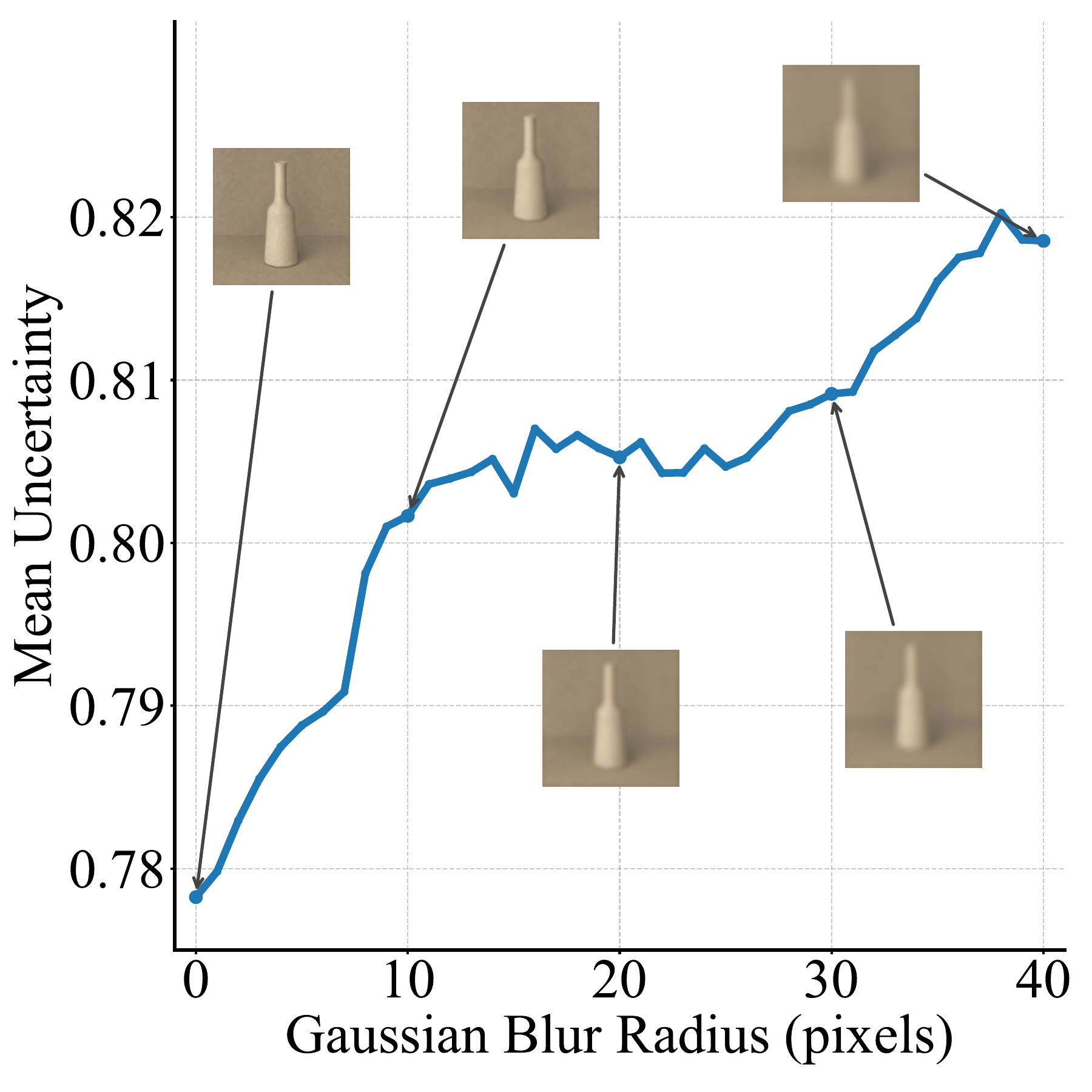}
        \textbf{(c)} Model Confidence
    \end{minipage}
    \caption{\textbf{Analysis of Matisse's predicted uncertainty.}
    (a) The pagoda region outside the input image's field of view exhibits
    higher uncertainty than visible regions.
    (b) Uncertainty remains similar at rotationally corresponding locations
    on the bottle. The concentric rings in the polar plot indicate approximate
    invariance across yaw angles, consistent with the object's rotational symmetry.
    (c) Higher mean uncertainty over occupied tokens is consistent with lower model
    confidence as Gaussian blur removes visual detail from the bottle image.}
    \label{fig:uncertainty-diagnostics}
\end{figure}

\myParagraph{Visibility}
Figure~\ref{fig:uncertainty-diagnostics}(a) illustrates how
visibility affects uncertainty. The input image captures only
part of the pagoda, with its lower-right region outside the
field of view. This unobserved region exhibits higher
uncertainty than the visible regions.

\myParagraph{Symmetry}
We examine whether the predicted uncertainty field reflects the rotational
symmetry of a bottle. We compare uncertainty at occupied token locations 
with that at their counterparts under yaw rotations about the vertical axis.
Figure~\ref{fig:uncertainty-diagnostics}(b) shows uncertainty patterns
across yaw angles. In the polar plot, the radial axis indexes fixed voxel
locations in 3D space, while the angular axis represents the yaw angle.
As the object rotates, different object voxels may occupy the same spatial
location. Because these voxels are related by the bottle's rotational symmetry,
their uncertainty values remain nearly constant across angles, producing
concentric rings in the plot. These results indicate approximate rotational 
symmetry of the predicted uncertainty field, with small directional differences.

\myParagraph{Model Confidence}
We examine whether Matisse's uncertainty reflects reduced model confidence
as visual detail is lost. We apply progressively stronger Gaussian blur to
the same bottle image and rerun the prediction pipeline with fixed inference
settings. As shown in Figure~\ref{fig:uncertainty-diagnostics}(c), mean
uncertainty over occupied tokens rises overall from $0.778$ for the original
image to $0.819$ at a blur radius of 40 pixels. Although the increase is not
strictly monotonic, the overall trend is consistent with reduced model
confidence as image quality degrades.

\section{Robustness to Repeated and Correlated Observations}
\label{app:repeated-views}
Our Evidential Uncertainty (EU) formulation assumes conditional independence 
across observations. Repeated or nearly identical views may violate this 
assumption, potentially leading to overestimated confidence. This is a 
common challenge in information-theoretic active reconstruction. 
For instance, FisherRF \citep{jiang2023eccv-fisherrf} also accumulates 
Fisher information additively, which may overestimate information under 
repeated observations. To mitigate redundant viewpoint selection, 
FisherRF employs greedy batch selection \citep[Section~3.3]{jiang2023eccv-fisherrf}, 
updating the information matrix after each selected candidate to encourage 
complementary views.

Matisse follows a similar principle by incorporating new observations 
through Equation~\eqref{eq:stream3d-evidence-fusion} and evaluating candidate 
viewpoints using updated EU, thereby reducing the expected information 
gain of already-observed regions. Our geometric exploration factor further 
discourages selecting nearby viewpoints. Moreover, as described in 
Section~\ref{sec:preliminary}, Adaptive Evidential Memory (AEM) retains 
at most the \(D=5\) highest evidence weights per token. This bounds 
evidence accumulation and prevents repeated observations from indefinitely 
increasing the retained evidence.

While these mechanisms mitigate redundant observations, they do not 
explicitly model cross-view correlations. We acknowledge this as a 
limitation of the conditional independence approximation.

\section{Extended Related Work}
\label{sec:appendix-related-work}

\myParagraph{Scene Representations for Active Reconstruction}
Scene representations have played a central role in active perception.
Early methods reason about geometric landmarks
\citep{sim2005icra-global} or probabilistic occupancy grids
\citep{bourgault2002iros-information,stachniss2005rss-information,
carlone2010iros-application,jadidi2015iros-mutual,bircher2016icra-receding,
zhou2021ral-fuel,guedon2022neurips-scone,guedon2023cvpr-macarons}.
While these representations support efficient visibility and information-gain
reasoning, their spatial resolution and memory consumption are constrained by
discretization.

More recently, implicit neural fields and Gaussian splatting have enabled
photorealistic rendering and high-fidelity novel-view synthesis.
Active reconstruction methods based on NeRFs
\citep{lee2022ral-uncertainty,pan2022eccv-activenerf,
yan2023iccv-active,feng2024cvpr-naruto}
and Gaussian splats
\citep{jiang2023eccv-fisherrf,xie2025rss-gauss,
li2025ral-activesplat,chen2025cvpr-activegamer,
jin2025ral-activegs,xue2026cvpr-uncertainty,
jeong2026icra-informative,jun2026eccv-sa}
use these representations to evaluate candidate observations.
However, their predictions are generally reliable only near observed regions,
and updating the scene representation often requires expensive iterative optimization
after acquiring new observations.

Several methods introduce learned completion priors to reason beyond current
observations \citep{guedon2022neurips-scone,guedon2023cvpr-macarons}.
Most closely related to our work, \citet{li2026cvpr-magician} predict occupancy
in unseen space, then convert those predictions into Gaussian primitives for uncertainty rendering,
thereby avoiding costly online optimization of a Gaussian field.
In contrast, we use a generative 3D foundation model as the reconstruction
backbone and maintain the scene as an implicit 3D evidence memory.
This formulation provides a learned geometric completion prior for occluded
and unobserved regions while permitting efficient uncertainty propagation as new observations
arrive.

\myParagraph{Uncertainty Modeling for Active Reconstruction}
Active perception requires an uncertainty measure that can predict the utility
of candidate observations. Early approaches maintain probabilistic occupancy
maps and select actions according to expected information gain
\citep{bourgault2002iros-information,stachniss2005rss-information, carlone2010iros-application,
jadidi2015iros-mutual,bircher2016icra-receding}. Learned occupancy-completion methods can additionally 
estimate uncertainty in unobserved regions \citep{guedon2022neurips-scone,guedon2023cvpr-macarons}.

More recently, radiance-field representations, including implicit NeRFs \citep{mildenhall2021acm-nerf}
and explicit 3D Gaussian splats \citep{kerbl2023acm-3d}, have enabled photorealistic rendering
and high-quality novel-view synthesis. Building on these representations,
recent neural reconstruction methods derive uncertainty from various
properties of the underlying fields and rendering process. 
NeRF-based methods usually learn an uncertainty field from reconstruction. \citet{pan2022eccv-activenerf} 
predict color variance; \citet{feng2024cvpr-naruto} maintain and update an explicit 3D
uncertainty field; and \citet{xue2024cvpr-neural} estimate uncertainty
from view-dependent transmittance and learn a spatially smooth uncertainty
field. Other methods estimate uncertainty from the behavior or information content of 
the reconstruction parameters. For example, \citet{yan2023iccv-active} estimate SDF variability under parameter
perturbations. Some methods derive uncertainty directly from quantities involved in the rendering equation.
\citet{lee2022ral-uncertainty} use the entropy of volumetric rendering weights.

Gaussian-splatting approaches commonly associate uncertainty or confidence
with individual primitives. This includes image-space rendering coverage
\citep{chen2025cvpr-activegamer}, combined 2D and 3D coverage
\citep{li2025ral-activesplat}, geometrically computed surfel confidence
\citep{jin2025ral-activegs}, measures based on transmittance and directional
view similarity \citep{xue2026cvpr-uncertainty}, and uncertainty primarily derived
from occupancy predictions \citep{li2026cvpr-magician}. These approaches use uncertainty 
to guide view selection. However, observing the most uncertain region does 
not necessarily yield the greatest information gain: the value of a candidate 
view depends on how much it is expected to reduce uncertainty, 
rather than on the current uncertainty alone.

Most closely related to our work, FisherRF \citep{jiang2023eccv-fisherrf} estimates parameter-space 
information from the curvature of the rendering likelihood and approximates the expected
information gain of candidate views. \citet{jun2026eccv-sa} combine this formulation 
with a feed-forward reconstruction prior. GauSS-MI
\citep{xie2025rss-gauss} takes a different information-theoretic approach: it
maintains a Bayesian reliability probability for each Gaussian based on
rendering residuals and selects views according to the Shannon mutual
information between Gaussian reliability and future observations.
GAVIS \citep{xue2026cvpr-uncertainty} introduces anisotropic visibility fields 
and a Bayesian network-based uncertainty-aware 3DGS rasterizer, 
selecting viewpoints by maximizing the predictive entropy of 
synthesized observations within an information-gain framework.

However, these formulations remain tied to the currently instantiated
scene representation. FisherRF relies on a local, second-order approximation
around the current parameter estimate, GauSS-MI models the reliability
of existing Gaussian primitives, and GAVIS estimates visibility-dependent 
uncertainty through Gaussian-splatting-based rendering. 
Consequently, they do not explicitly
represent a distribution over geometries that may appear in the future.

In contrast, we derive uncertainty and information gain from plausible scene completions
induced by a diffusion-based generative model. Our formulation is therefore global in scene-hypothesis
space, rather than being restricted to a local geometry around the
current model parameters. Moreover,
the diffusion prior allows the information-gain objective to account for
structural predictability, such as symmetry and geometric regularity, and model confidence, such as 
out-of-distribution uncertainty.

\myParagraph{Uncertainty in Generative Models}
Uncertainty estimation for generative models generally follows four
strategies: training an auxiliary uncertainty predictor
\citep{horwitz2022arxiv-conffusion}, constructing an ensemble of independently
trained models \citep{berry2024uai-shedding}, estimating a posterior
distribution over model parameters
\citep{kou2024iclr-bayesdiff,jazbec2025arxiv-generative}, or querying an
external oracle as an uncertainty evaluator
\citep{franchi2025cvpr-towards}. These approaches, however, were developed
primarily outside 3D shape reconstruction and do not directly address
cross-view uncertainty consistency.

In 3D reconstruction, RIGI \citep{wang2026tip-rigi} estimates uncertainty in image
space and incorporates an uncertainty-aware objective during training, while
LatentSplat \citep{wewer2024eccv-latentsplat} explicitly predicts the standard
deviation of its outputs. Both approaches require uncertainty-specific model
training. ProbNeRF \citep{hoffman2023icais-probnerf} instead estimates
pixel-wise uncertainty at inference time by repeatedly sampling latent codes
and model weights using Hamiltonian Monte Carlo.
\citet{tewari2023diffusion} similarly estimate pixel-wise uncertainty from
multiple generations initialized with different noise samples. Because these
methods define uncertainty separately in each rendered image, evaluating a
new camera view requires multiple additional model evaluations, making
uncertainty estimation costly for active perception.

More recently, \citet{xiang2026arxiv-dvd} estimate occupancy-grid entropy at
inference time, but do not aggregate uncertainty consistently across
multiple observations and viewpoints. In contrast, our method extracts
uncertainty from a pretrained 3D generative foundation model without
uncertainty-specific training and fuses it across observations in a shared
3D latent representation. Once constructed, this representation supports
consistent and efficient uncertainty evaluation from multiple candidate
views. To the best of our knowledge, we are the first to introduce
training-free, multi-view uncertainty estimation for a foundational 3D
generative model.

\section{Limitations}
\label{app:limitations}

Matisse builds on an underlying pretrained 3D generator and the fusion mechanism of
Stream3D \citep{zhou26arxiv-stream3d}. Consequently, its reconstruction quality is 
limited by the learned generator prior and by the ability of Stream3D to 
associate and fuse observations accurately. Failures in
generating plausible geometry, i.e., sparse-structure stage output, 
or fusing inconsistent observations
can propagate to the uncertainty estimates produced 
by Matisse and, in turn, to its view-selection and
keyframe-retention decisions.
\section{Future Work}
\label{app:future_work}
We conclude by highlighting several promising directions for future research.

First, recent multi-view diffusion models
\citep{zadaianchuk2026reconstruction-eccv,wu2026gator-arxiv}
share architectural similarities with SAM3D, raising the question
of whether our EU formulation generalizes
to these architectures. Similarly, GenRecon
\citep{schmid2026genrecon} extends generative reconstruction
to scene-level settings, motivating further investigation into
the applicability of our EU and evidential information gain
(EIG) formulations to scene-level generation.

Second, while Matisse focuses on reducing geometric uncertainty,
extending EIG to task-aware active perception
\citep{Maggio24ral-clio} could enable robots to prioritize
observations that directly support downstream manipulation
and planning.

Third, Matisse primarily considers geometric uncertainty
without explicitly reasoning about physical consistency.
In cluttered scenes, ambiguities in object geometry, pose,
and contact relationships can lead to physically implausible
reconstructions. Incorporating physical constraints, such as
non-penetration and static stability
\citep{yu2026picasso,huang2026simulation,ma2026rest3d,ranawaka2026simfoundry},
could not only improve reconstruction plausibility but also
guide active viewpoint selection toward regions where
additional observations are most valuable for resolving
physical inconsistencies.

Finally, our EU formulation assumes conditional independence
across observations, potentially leading to overconfidence
under highly correlated views. Incorporating correlation-aware
evidence fusion could improve uncertainty calibration
and robustness to redundant observations.

\end{document}